%% file: aspfoa-tr.tex
\documentclass[oneside,a4paper]{article}

\newif\ifdraft\draftfalse
\newif\ifappendix\appendixtrue
\newif\ifdotikz\dotikzfalse
\dotikztrue
\newif\ifrevisionmarkers\revisionmarkersfalse
\usepackage{amssymb} %
\usepackage{amsthm}
\usepackage{amsmath}
\usepackage{url}
\usepackage{paralist}
\usepackage{booktabs}
\usepackage{multicol}
\usepackage{multirow}
\usepackage[sort,numbers]{natbib}
\usepackage{fullpage}
\usepackage{graphicx}
\usepackage[normalem]{ulem}
\usepackage[T1]{fontenc} %
\usepackage[vlined,algoruled,linesnumbered]{algorithm2e}

\ifdraft
\usepackage{todonotes}
\newcommand{\comment}[1]{\todo[inline,color=orange!40]{#1}}
\else
\newcommand{\comment}[1]{}
\fi

\ifrevisionmarkers
\usepackage{color}
\newcommand{\rev}[2]{{\color{red}#1}{\color{blue}#2}}
\else
\newcommand{\rev}[2]{#2}
\fi

\ifdotikz
\usepackage{tikz}
\usetikzlibrary{shapes,arrows,backgrounds,%
matrix,patterns,arrows,decorations.pathmorphing,decorations.pathreplacing,%
positioning,fit,calc,decorations.text,shadows%
}
\pgfrealjobname{aspfoa-tr}
\else
\long\def\beginpgfgraphicnamed#1#2\endpgfgraphicnamed{\includegraphics{#1}}
\fi

\usepackage{mymacros}
\usepackage{myfiguremacros}

\begin{document}
\title{%
{\sc Technical Report}\\[1em]
Modeling Variations of First-Order Horn Abduction \\
in Answer Set Programming}

\author{Peter Sch\"uller%
\thanks{This work is a significant extension of \cite{Schuller2015rcra}; major additions are preference relations \accelobj\ and \waobj, revised encodings, increase performance, on-demand constraints, and flexible value invention.
This work has been supported by Scientific and Technological Research Council of Turkey (TUBITAK) Grant 114E777.
This document is a preprint of \cite{Schuller2016aspfoa} with minor formatting corrections.}%
\\
Institut f\"ur Logic and Computation, Knowledge-Based Systems Group \\
Technische Universit\"at Wien, Austria \\
Computer Engineering Department, Faculty of Engineering\\
Marmara University, Turkey \\
{\tt schueller.p@gmail.com}}

\maketitle

\begin{abstract}
\input{abstract.txt}
\end{abstract}

\ifdraft
\listoftodos
\fi

\section{Introduction}

Abduction \cite{Peirce1955} is reasoning to the best explanation,
which is an important topic in diverse areas
such as diagnosis, planning, and natural language understanding (NLU).
\rev{%

We here focus on a variant of abduction used in NLU
where the primary concern is to find a preferred explanation of a given input
(sentence) with respect to an objective function,
knowledge is expressed in First Order (FO) Horn logic,
and the Unique Names Assumption (UNA) does not hold in general.
In this reasoning problem,
the space of potential explanations can be infinite,
existing tools realize different solutions to make reasoning decidable,
and these solutions are ad-hoc and not formally described.
Moreover, several objective functions exist,
and they are realized by different tools which makes their comparison difficult.
Also, global consistency constraints are necessary to yield practically meaningful solutions,
however existing solvers realize each possible form of a global constraint (e.g., unique slot values)
in a separate checking procedure and make it difficult to experiment with additional constraints.

Towards overcoming some of these problems,
we realize abduction in the declarative formalism of
Answer Set Programming (ASP) \cite{Lifschitz2008}
which allows modular modeling of combinatorial problems
based on clearly defined formal semantics.
Primary motivation for this work is to obtain a more flexible framework
where variations of objective functions and global constraints on abduction can be studied easily,
for example to study novel preferences such as \cite{Schuller2014winograd}
that can comfortably be represented in ASP but not in other solvers.
A secondary motivation is to use ASP for realizing a task that it is not typically used for,
and to study how far we can go in this task
without built-in reasoner support for term equivalence and existential variables.
While pursuing these motivations,
we obtain insights on the structure of the problem and several objective functions,
as well as insights on the efficiency of straightforward versus more involved ASP formulations.

We compare our approach experimentally with the state-of-the-art solver
\phillip\ \cite{Yamamoto2015} and find that our approach
is significantly faster for instantiating and solving the problem
to the optimum result.

The concrete preference relations we study in this work are
Weighted Abduction \cite{Hobbs1993},
Coherence \cite{Ng1992kr}, and cardinality minimality.
These objective functions are based on a proof graph,
induced by back-chaining and unification operations.

Representing back-chaining in ASP
requires existential variables in the head of rules,
which is not directly supported in ASP,
and the standard solutions of using uninterpreted function symbols
do not guarantee a finite instantiation.
Moreover, unification and absence of the UNA is an additional challenge,
because ASP semantics is based on the UNA.

We describe how we tackled the above challenges
and make the following contributions.
}{

We here focus on a variant of abduction, used in NLU,
where the primary concern is to find an explanation of a given input
(sentence) with respect to an objective function.
Knowledge is expressed in First Order (FO) Horn logic axioms.
For example \quo{a father of somebody is male} can be expressed as follows, where capital letters are variables
which are universally quantified unless explicitly indicated otherwise.
\begin{align*}
  \miinst(X,\mimale) \lax \mifatherof(X,Y).
\end{align*}
Abduction aims to find a set of explanatory atoms
that make a set of goal atoms true with respect to a
background theory (i.e., a set of axioms).
If $\miinst(\mi{tom},\mimale)$ is part of a goal
then abduction can explain this goal atom with the atom
$\mifatherof(\mi{tom},\mimary)$
where $\mimary$ is another person of interest.
Using abduction,
we can interpret whole natural language texts,
for example
\quo{Mary lost her father. She is depressed.}
can be interpreted using knowledge about
losing a person, death of a person, and being depressed,
such that we obtain an abductive explanation
that represents
\quo{Mary's father died, and this is the reason for her depression}.

Abductive reasoning in FO Horn logic
yields an infinite space of potential inferences,
because backward reasoning over axioms
can produce existentially quantified variables,
which can introduce new terms (value invention).
For example the above axiom is transformed as follows.
\begin{align*}
  \miinst(X,\mimale) \Rightarrow \exists Y: \mifatherof(X,Y).
\end{align*}
To achieve decidability,
we need to limit value invention,
which leads to a challenging trade-off: the more we limit value invention,
the more (potentially optimal) solutions we lose.

A second challenge in FO logic is,
that terms (input and invented) can be equivalent to other terms.
Equivalent terms make distinct atoms equivalent,
which is used in an inference called \emph{factoring}.
In the above example, we can say that
$\miis(\mimary,\midepressed)$
is factored with
$\miis(\mi{she},\midepressed)$
under the assumption that the equivalence
$\mimary \eqs \mi{she}$ holds.

A crucial issue when using abduction for NLU is the choice
of an appropriate \emph{preference} among possible
abductive explanations.
\emph{Cardinality minimality} of the set of abduced atoms
is a frequently used preference,
however in NLU two other preferences have turned out to be more effective:
\emph{coherence} \cite{Ng1990,Ng1992thesis},
and \emph{weighted abduction} \cite{Stickel1989,Hobbs1993,Singla2011},
which are based on a proof graph induced by
back-chaining and unification operations.

Several tools for realizing abduction with such preferences
exist: \phillip,
based on Integer Linear Programming (ILP) \cite{Yamamoto2015}
and its precursor \henry~\cite{Inoue2013}
as well as an approach based on Markov Logic \cite{Blythe2011}.
The problem of termination is solved in \cite{Blythe2011}
by instantiating existential terms only with terms
from the input (no value invention),
while \cite{Inoue2013,Yamamoto2015} solves
the issue by inventing a new term
only if no previously invented term is present
in the head of the axiom.

Unfortunately, using only input terms eliminates
many valid solutions in NLU applications,
for example when processing a text
about a son and a grandfather, we would be unable to perform
reasoning about the father
(because it does not exist as a constant).
The alternative approach of blocking value invention if an
invented value is involved in the rule improves the situation,
however it is an ad hoc solution
and its implications on solution quality
have not been formally or experimentally analyzed.

In addition to decidability issues, global consistency constraints are necessary to yield practically meaningful abductive explanations,
however existing solvers \henry\ and \phillip\ realize each possible form of a global constraint (e.g., unique slot values)
in a separate checking procedure and make it difficult to experiment with additional constraints.

Towards overcoming some of these problems,
we realize abduction in the declarative formalism of
Answer Set Programming (ASP) \cite{Lifschitz2008}
which allows modular modeling of combinatorial problems
based on clearly defined formal semantics.
Primary motivation for this work is to obtain a more flexible framework
where variations of \rev{}{Skolemization, }objective functions and global constraints on abduction can be studied easily
and where novel preferences can be studied,
such as \cite{Schuller2014winograd}
that can comfortably be represented in ASP but not in other solvers.
Our secondary motivation is,
to use ASP for realizing a task that it is not typically used for,
and to study how far we can go in this task.

Realizing FO Horn abduction in ASP
poses further challenges for the following reasons:%
\begin{inparaenum}[(i)]
\item
  ASP semantics is based on the
  Unique Names Assumption (UNA),
  which means that distinct terms cannot be equivalent
  (e.g., $\mi{mary} = \mi{she}$ is not expressible
  in a built-in feature of ASP); moreover
\item
  ASP rules have no built-in support for existential
  variables in rule heads, which is necessary for
  value invention during back-chaining as shown above.
  In particular, Skolemization using function symbols,
  i.e., replacing
  $\exists Y: \mifatherof(X,Y)$ by
  $\mifatherof(X,sk(X))$ where $sk$ is a new function symbol,
  does not guarantee a finite instantiation.
\end{inparaenum}

We tackle the above challenges
and present an ASP framework for solving
FO Horn abduction problems
for objective functions
weighted abduction, coherence, and cardinality minimality.
We describe insights on the structure of the problem
as well as insights on the efficiency
of straightforward versus more involved ASP formulations.
Our formulation allows a fine-grained configuration of Skolemization for tackling cyclic background theories,
moreover it permits the usage of global constraints
of any form that is expressible in ASP.
Experiments show, that our framework is
faster than the state-of-the-art solver \phillip\ \cite{Yamamoto2015}
on the \accel\ benchmark \cite{Ng1992kr}
for plan recognition in NLU.
\smallskip

In detail, we make the following contributions.
}
\begin{itemize}
\item
  \rev{}{We provide a novel uniform formalization of abduction
  with preference relations weighted abduction,
  coherence, and cardinality minimality
  (Section~\ref{secPrelims}).}
\item
  We \rev{}{present an ASP encoding that realizes
  back-chaining in ASP}
  by deterministically represent\rev{}{ing an}
  abductive proof graph\rev{s in ASP encoding \bwd, based on}%
  { and guessing which parts of that graph to use.}
  For value invention we use uninterpreted function terms,
  and we \rev{}{explicitly} represent an equivalence relation
  between terms to model unification
  \rev{}{(Section~\ref{secBwd})}.
\item
  We describe canonicalization operations on proof graphs,
  show that \rev{these transformations}{they} do not eliminate optimal solutions,
  and \rev{use these properties to encode unification
  more efficiently}{use these transformations for
  encoding factoring efficiently
  (Section~\ref{secFactoring})}.
\item
  We \rev{describe another}{present an alternative} ASP encoding\rev{ (\fwda)}{}
  which does not represent a proof graph,
  instead it generates abduced atoms, defines truth using axioms,
  and tests if goals are reproduced
  \rev{}{(Section~\ref{secFwda})}.
\item
  We give ASP encodings for \rev{}{realizing the }objective functions \rev{for }{}weighted abduction,
  coherence, and cardinality minimality%
  \rev{}{\ (Section~\ref{secPreferences})}.
\item
  We \rev{extend our}{study an alternative} method for value invention
  by replacing uninterpreted function terms with external computations\rev{ in Python that allow}%
  {. This provides us with} a fine-grained control over Skolemization,
  which is more flexible than state-of-the-art solutions
  for achieving decidability.
  We formalize this extension using the \hex\ formalism,
  \rev{which allows us to}{and} show termination guarantees for arbitrary
  (i.e., including cyclic) knowledge bases%
  \rev{}{ (Section \ref{secFlexibleValueInvention})}.
\item
  We apply a technique used to increase performance of the \henry\ solver
  \rev{}{\cite{Inoue2013} }for weighted abduction \rev{\cite{Inoue2012}}{}%
  to our encodings by introducing on-demand constraints
  for transitivity of the equivalence relation
  and for \rev{a relation used in formulating}{ensuring} acyclicity of the proof graph.
  We \rev{also }{}formalize this using \hex\ and describe an algorithm for computing optimal models
  in the presence of on-demand constraints using the Python API of \rev{the }{}\clingo\rev{ solver}{}~%
  \cite{Gebser2014clingorepext} \rev{}{(Section \ref{secOnDemand})}.
\item
  We perform computational experiments on the \accel\ benchmark \cite{Ng1992kr}%
  \rev{ for plan recognition in NLU ,
  describe how to translate global constraints of \accel\ into ASP.
  We}{ where we} measure and discuss resource consumption in terms of space, time, and solver-internal statistics,
  as well as solution quality in terms of the objective function
  \rev{}{(Section~\ref{secExperiments})}.
  For experimental evaluation we use
  \rev{}{the Python API of \clingo},
  \gringo\ \rev{}{\cite{Gebser2011gringo3} }with
  \rev{}{either }\clasp\ \rev{}{\cite{Gebser2015clasp3} }or
  \wasp\ \rev{}{\cite{Alviano2015wasp}},
  and the \rev{existing }{\phillip\ \cite{Yamamoto2015} }%
  solver for weighted abduction\rev{ \phillip}{}
  which is based on C++ and Integer Linear Programming (ILP).
  \rev{\\ Our framework, including experimental instances and algorithms,
is publicly available.}{}
\end{itemize}
\rev{}{%
We discuss related work in Section~\ref{secRelated}
and conclude in Section~\ref{secConclusion}.

Appendices provide additional information:
all ASP encodings in their complete version;
verbose listing of rewriting and answer set of the running example;
proofs for correctness of encodings}%
\rev{Proofs for encodings are given as sketches
in Appendix~\ref{secProofs} and intended to
illustrate why each rule of our encodings
is required and why certain assumptions
and details of our formalizations are necessary}%
{; and algorithms for realizing on-demand constraints}.

\rev{}{Our framework, including experimental instances and algorithms,
is available online.}%
\footnote{\url{https://bitbucket.org/knowlp/asp-fo-abduction}}

\section{Preliminaries}
\label{secPrelims}

We give a brief introduction of \rev{A}{a}bduction in general
and variations of First Order Horn abduction
as used in Natural Language Processing,
describe the \accel\ benchmark which contains instances of such reasoning problems,
and give preliminaries of ASP and HEX.

In the following, in logical expressions and ASP rules
we write variables starting with capital letters
and constants starting with small letters.

\subsection{Abduction and Preferences on Abductive Explanations}
\label{secPrelimsAbduction}

Abduction, originally described in \cite{Peirce1955},
can be defined logically as follows:
given a set $B$ of background knowledge axioms
and an observation $O$,
find a set $H$ of hypothesis atoms
such that $B$ and $H$ are consistent
and reproduce the observation,
i.e., $B \cups H \not\models \bot$
and $B \cups H \models O$.
In this work we formalize axioms and observations
in First Order logic:
the observation $O$ (also called \quo{goal})
is an existentially quantified conjunction of atoms
\begin{equation}
  \label{eqObs}
  \exists V_1,\ldots,V_k: o_1(V_1,\ldots,V_k) \lands \cdots \lands o_m(V_1,\ldots,V_k)
\end{equation}
and an axiom in $B$ is a Horn clause of form
\begin{equation}
  \label{eqAx}
  q(Y_1,\ldots,Y_m) \,\lax\, p_1(X^1_1,\ldots,X^1_{k_1}) \lands \cdots \lands p_r(X^r_1,\ldots,X^r_{k_r}).
\end{equation}
where $\cX = \bigcup\rev{}{_{1 \les i \les r}}\rev{}{\bigcup_{1 \les j \les k_r}} X^i_j$ is the set of variables in the body,
$\cY = \bigcup\rev{}{_{1 \les i \les m}} Y_i$ is the set of variables in the head,
$\cY \subseteqs \cX\strut$ and we implicitly quantify universally over $\cX$.
In the variant of abduction we consider here,
the set $H$ of hypotheses can contain any predicate
from the theory $B$ and the goal $O$,
hence existence of a solution is trivial.
A subset $S$ of constants from $B$
is declared as \quo{sort names} that cannot be
equivalent with other constants.
Given $B$, $O$, and $S$,
we call the tuple $(B,O,S)$
an \emph{abduction instance}.
Unless otherwise indicated, we assume that $B$ is acyclic.

\begin{example}[Running Example]
\label{mainexample}
Consider the following text
\medskip

\centerline{\quo{\emph{Mary lost her father. She is depressed.}}}
\medskip

\noindent
which can be encoded as the following observation,
to be explained by abduction.
\begin{eqnarray}
  \label{eqExGoal}
  \miname(m,\mimary) {\wedge} \milost(m,f) {\wedge}
  \mifatherof(f,m) {\wedge} \miinst(s,\mifemale)
  {\wedge} \miis(s,\midepressed)
\end{eqnarray}
Given the set of axioms
\begin{eqnarray}
  \label{eqEx1}
  \miinst(X,\mimale) &\lax& \mifatherof(X,Y) \\
  \label{eqEx2}
  \miinst(X,\mifemale) &\lax& \miname(X,\mimary) \\
  \label{eqEx5}
  \!\!\!\!\miimportantfor(Y,X) &\lax& \mifatherof(Y,X) \\
  \label{eqEx7}
  \miinst(X,\miperson) &\lax& \miinst(X,\mimale) \\
  \label{eqEx3}
  \miis(X,\midepressed) &\lax& \miinst(X,\mipessimist) \\
  \label{eqEx4}
  \miis(X,\midepressed) &\lax& \miis(Y,\midead) \lands \miimportantfor(Y,X) \\
  \label{eqEx6}
  \milost(X,Y) &\lax& \miis(Y,\midead) \lands \miimportantfor(Y,X) \lands \miinst(Y,\miperson)
\end{eqnarray}
and sort names
\begin{eqnarray}
  \miperson \quad \mimale \quad \mifemale \quad \midead \quad \midepressed
\end{eqnarray}
we can use abduction to conclude the following:
(a) loss of a person here should be interpreted as death,
(b) \quo{she} refers to Mary,
and (c) her depression is because of her father's death
because her father was important for her.

We obtain these because we can explain \eqref{eqExGoal}
by the following abductive explanation
which contains atoms and equivalences.
\begin{eqnarray}
  \label{eqAbduced}
  \miname(m,\mimary) \quad \mifatherof(f,m) \quad \miis(f,\midead) \quad m = s
\end{eqnarray}
The first two atoms directly explain goal atoms.
We can explain the remaining goal atoms
using inference from rules and
\rev{unification (also called \quo{factoring})}%
{factoring (which represents unification in a certain reasoning direction)}.
\begin{eqnarray}
  \label{eqExfrom1}
  \miinst(f,\mimale)    &\quad& \text{[infered via \eqref{eqEx1}
    using \eqref{eqAbduced}]} \\
  \label{eqExfrom2}
  \miinst(m,\mifemale)  &\quad& \text{[infered via \eqref{eqEx2}
    using \eqref{eqAbduced}]}\\
  \miinst(s,\mifemale)  &\quad& \text{[goal,
    factored from \eqref{eqExfrom2} using \eqref{eqAbduced}]} \notag \\
  \label{eqExfrom5}
  \miimportantfor(f,m)  &\quad& \text{[infered via \eqref{eqEx5}
    using \eqref{eqAbduced}]} \\
  \label{eqExfrom7}
  \miinst(f,\miperson)  &\quad& \text{[infered via \eqref{eqEx7}
    using \eqref{eqExfrom1}]} \\
  \label{eqExfrom4}
  \miis(m,\midepressed) &\quad& \text{[infered via \eqref{eqEx4}
    using \eqref{eqAbduced} and \eqref{eqExfrom5}]} \\
  \miis(s,\midepressed) &\quad& \text{[goal,
    factored from \eqref{eqExfrom4} using \eqref{eqAbduced}]}  \notag \\
  \label{eqExfrom6}
  \milost(m,f)          &\quad& \text{[goal, infered via \eqref{eqEx6}
    using \eqref{eqAbduced}, \eqref{eqExfrom5}, and \eqref{eqExfrom7}]}
\end{eqnarray}
Note that there are additional possible inferences
but they are not necessary to explain the goal atoms.
Moreover, there are several other abductive explanations,
for example to abduce all goal atoms,
or to abduce $\miinst(m,\mipessimist)$ and $\milost(m,f)$
instead of abducing $\miis(f,\midead)$.
\hfill$\square$
\end{example}

\leanparagraph{Preferred explanations.}
In the presence of multiple possible explanations,
we are naturally interested in obtaining a set of
\emph{preferred} explanations.
In this work we consider three preference formulations:
\bipe{(a)}
\item[(\cardobj)]
  cardinality-minimality of abduced atoms,
\item[(\accelobj)]
  \quo{coherence} as described in \cite{Ng1992kr}
  and in slight variation in more detail in \cite{Ng1990}, and
\item[(\waobj)]
  \quo{weighted abduction} as initially formulated in \cite{Hobbs1993}.
\eipe

\accelobj\ is based on connectedness between observations and explanations,
while
\waobj\ finds a trade-off between least-specific and most-specific abduction,
depending on the explanatory power of more specific atoms.
Both objective functions are based on an inference procedure
that induces a proof graph by means of backchaining and unification.

Towards formalizing these preference functions,
we next formalize such an inference procedure.
The following definitions are based on \cite{Stickel1989,Inoue2014rte}
but additionally define an explicit proof graph
corresponding to the inferences leading to a hypothesis.
Here we consider only single-head axioms.
\begin{definition}
  \label{defHypotheses}
  A \emph{hypothesis} is a conjunction of atoms or equivalences between terms.
  Given an abduction instance $A \eqs (B,O,S)$
  the set $\hcH(A)$ of all hypotheses of $A$ is the largest set
  containing hypotheses obtained by extending
  $\cH \eqs \rev{\emptyset}{\{O\}}$
  using back-chaining and unification.

  \rev{}{\emph{Back-chaining}:}
  given an atom $P$ which is \rev{either an observation ($P\ins O$) or}{}
  part of a hypothesis ($P \ins H$, $H \ins \cH$),
  such that $P$ unifies with the head $Q \eqs q(Y_1,\ldots,Y_m)$
  of an axiom $\eqref{eqAx}$ with substitution $\theta$,
  back-chaining adds to $\cH$
  a new hypothesis $H \lands P_1' \lands \cdots \lands P_r'$,
  where $P_i'$  is the substituted version
  $\theta(p_i(X^i_1,\ldots,X^i_{k_i})$
  of the $i$-th body atom of axiom \eqref{eqAx}.

  \rev{}{\emph{Unification}:}
  given \rev{two }{hypothesis $H \ins \cH$ with distinct} atoms $P,Q \ins H$\rev{, $H \ins \cH$,}{}
  that unify under substitution $\theta$
  such that all $X \,{\mapsto}\, Y \ins \theta$ obey $X,Y \notins S$,
  unification adds hypothesis
  $H {\wedge} \bigwedge \{ X \eqs Y \mids X \,{\mapsto}\, Y \ins \theta \}$ to $\cH$.
\end{definition}
Note that $\hcH(A)$ is potentially infinite. We sometimes leave $A$ implicit.
\begin{example}[continued]
  \rev{}{
  Three hypotheses for the abduction instance in Example~\ref{mainexample}
  are%
  \begin{inparaenum}[(a)]%
  \item%
    the original goal $G$ as shown in \eqref{eqExGoal},
    which intuitively means that we do not justify any atom in the goal
    by inference, instead we abduce all atoms in the goal;
  \item
    the hypothesis
    $G
      \wedge \miis(f,\midead) \wedge m \eqs s
      \wedge \miinst(f,\mimale) \wedge \miinst(m,\mifemale)
      \wedge \miimportantfor(f,m) \wedge \miinst(f,\miperson)
      \wedge \miis(m,\midepressed)$
    which corresponds to \eqref{eqExfrom1}--\eqref{eqExfrom6}
    and includes the abductive explanation \eqref{eqAbduced}; and
  \item
    the hypothesis
    $G
      \wedge \miname(s,\mimary) \wedge m \eqs s
      \wedge \miinst(f,\miperson) \wedge \miinst(f,\mimale)
      \wedge \mifatherof(f,n_2) \wedge m \eqs n_2
      \wedge \miis(n_1,\midead) \wedge f \eqs n_1 \wedge \miis(f,\midead)
      \wedge \miimportantfor(f,m)
      \wedge \miimportantfor(n_1,m)$
    which applies Skolemization during back-chaining
    and represents a variation of explanation \eqref{eqAbduced}.
    Details about this hypothesis can be found in Example~\ref{exWAGraph}
    and in Figure~\ref{figExWA}.
  \end{inparaenum}
  \hfill$\square$
  }
\end{example}

A hypothesis $H \ins \hcH$ does not contain any information
about how it was generated.
For the purpose of describing the cost function formally,
we \rev{formally }{}define proof graphs $G$ wrt.\ hypotheses $H$.
\begin{definition}
  \label{defHypGraph}
  Given an abduction instance $A \eqs (B,O,S)$,
  a \emph{proof graph}\ $G$ wrt.\ a hypothesis $H \!\ins \rev{\cH}{\hcH}(A)$
  is an acyclic directed graph consisting of nodes
  $N(G) \eqs \rev{O \cups}{}\{ \rev{H \mids H}{P \ins H \mids P} \text{ is not an equality} \}$
  and edges $E(G)$ are recursively defined
  by the inference operations used to generate \rev{$H$}{atoms $P \ins H$}:%
  \bipe{(a)}%
  \item back-chaining of $P$ induces an edge from all body atoms $P_i'$ to $P$, and
  \item unification of $P$ with $Q$ induces \rev{}{either }an edge from $P$ to $Q$\rev{}{\ or an edge from $Q$ to $P$.}
  \eipe
\end{definition}
We denote by
$A(G) \eqs \{ a \ins N(G) \mids \nexists b: (b,a) \ins E(G) \}$
the set of nodes that are neither back-chained nor unified.
\rev{}{Note that the term `factoring' is used to denote
unification with direction, this is discussed in detail
in \cite{Stickel1989}.
There can be multiple proof graphs with respect to a single hypothesis,
and these graphs differ only by factoring directions.
Figure~\ref{figExWA} depicts a proof graph
which is discussed in Example~\ref{exWAGraph}.
}

Intuitively, an edge in the proof graph
shows how an atom is justified:
by inference over an axiom (backchaining)
or by equivalence with another atom (\rev{unification}{factoring}).

Equipped with these definitions,
we next formalize the objective functions of interest.

\begin{definition}
  \label{defGoalFun}
  Given a proof graph $G$ wrt.\ a hypothesis $H$ of an abduction instance $(B,O,S)$,
  \begin{compactitem}
  \item
    $\cardobj \eqs |A(G)|$,
  \item
    $\accelobj \eqs |\{ (a,b) \mids a, b \ins O, a \lts b$,
    and $\nexists n \ins N(G)$
    such that from $n$ we can reach both $a$ and $b$ in $G\}|$
    \rev{}{where the relation $\lts$ is an arbitrary fixed total order over $O$ (e.g., lexicographic order)},
  \item
    $\waobj \eqs \sum_{a \ins A(G)} \min \micost(a)$,
    where $\micost\rev{}{\ {:}\ N(G) \to 2^\bbR}$ labels \rev{hypotheses }{each atom in the graph }with \rev{}{a set of }cost values.
  \end{compactitem}

  \noindent
    For \rev{}{the definition of $\micost$} in \waobj\ we require that each axioms of form \eqref{eqAx}
    has weights $w_1,\ldots,w_r$ corresponding to its body atoms,
    and initial costs $ic(o)$ for each observation $o \ins O$.

    Then $\micost$ is initialized with $\emptyset$ for each node
    and recursively defined as follows:
    \begin{compactitem}
    \item
      goal nodes $o \ins O$
      obtain cost $\micost(o) \eqs \micost(o) \cups \{ ic(o) \}$;
    \item
      if $P$ was back-chained with an axiom with body atoms $P_1',\ldots,P_r'$
      and $c \rev{\ins}{\eqs \min\,} \micost(P)$,
      then $c$ is added to each body atom $P_i'$ after
      adjusting it using the respective cost multiplier\rev{s }{ }$w_i$%
      \rev{:}{, formally}
      $\micost(P_i') \eqs \micost(P_i') \cups \{ c\rev{}{\,{\cdot}\,}w_i \}$
      for $1 \les i \les r$;
    \item
      if $P$ was unified with $Q$
      \rev{}{such that there is a factoring edge
      $(Q,P) \ins G$ from $Q$ to $P$,
      then }we add the smallest cost
      \rev{$c \eqs \min (\micost(P) \cups \micost(Q))$ to both $P$ and $Q$}{at $P$ to $Q$}:%
      \rev{ $\micost(P) \eqs \micost(P) \cups \{ c \}$ and }{}
      $\micost(Q) \eqs \micost(Q) \cups \{ \rev{c}{\min \micost(P)} \}$.
    \end{compactitem}
\end{definition}
Note that the formalization in \cite[above (1)]{Inoue2014rte}
assigns unification cost to the equality,
but does not use that cost in case of multiple unifications,
hence we do not use such a formalization.
Moreover, deleting cost values from the hypothesis with higher cost
in a unification (as shown in \cite[Fig.~1, `Output' vs `Backward-chaining']{Inoue2014rte})
contradicts the formalization as a cost \quo{function} that maps hypotheses to costs.
Therefore our formalization defines $\micost$ to map from atoms in a hypothesis
to multiple \quo{potential} costs of that hypothesis.
Note that due to acyclicity of the graph,
no cost in $\micost$ recursively depends on itself,
and that back-chaining can create hypothesis atoms containing a part of the goal,
therefore goal nodes can have a cost lower than that goal's initial cost
(e.g., $\mifatherof(f,m)$ in Figure~\ref{figExWA}).

\myfigureExWA{tb}
The formalization in this section
was done to provide a basis for showing correctness of canonicalization operations
on proof graphs and for showing correctness of ASP encodings.
To the best of our knowledge, no precise formal description of proof graphs
and associated costs of \accelobj\ and \waobj\ exists in the literature,
therefore we here attempt to formally capture the existing
descriptions
\cite{Ng1992kr,Hobbs1993,Inoue2013}.

\begin{example}[continued]
  \label{exWAGraph}
  The proof graph of Example~\ref{mainexample}
  is depicted in Figure~\ref{figExWA}
  where we also show the \rev{minimal cost }{set of costs }of each node
  using objective \waobj.
  The total cost of this graph is
  $100\$\pluss 48\$\pluss 40\$\eqs188\$$.
  Objective \cardobj\ has cost $3$ because we abduce 3 atoms,
  and objective \accelobj\ has cost $6$:
  let goal node set
  $A \eqs \{ \miinst(s,\mifemale)$, $\miname(m,\mimary) \}$
  and
  $B \eqs \{ \mifatherof(f,m)$, $\milost(m,f)$, $\miis(s,\midepressed) \}$,
  then nodes within $A$ and within $B$ are reachable from some node,
  however pairs $\{ (a,b) \mids a \ins A, b \ins B\}$ of nodes
  are not reachable from any node,
  and each of these $|A|\cdot|B| \eqs 6$ pairs incurs cost 1.
  \hfill$\square$
\end{example}

Note that in this work we consider only
hypotheses and proof graphs where an atom
is either justified by a single inference,
or by a single factoring, or not at all (then it is abduced).

\paragraph{Computational Complexity.}
Existence of an abductive explanation is trivial,
because we can abduce any atom and the goal is the trivial explanation.
However,
finding the optimal abductive explanation
with respect to an objective function is not trivial.
With unlimited value invention and cyclic theories
the problem of finding the optimal explanation is undecidable,
as we cannot bound the size of the proof graph
or the number of additionally required constants
for finding the optimal solution.

To the best of our knowledge,
the computational complexity of deciding whether an abductive explanation
is optimal wrt.\ one of the objective functions
\cardobj, \accelobj, and \waobj,
has not been formally studied so far,
although for \cardobj\ related results exist.
Section~\ref{secRelated} discusses related complexity results.

\subsection{\accel\ Benchmark}
\label{secAccel}
The \accel\ benchmark%
\footnote{Available at \url{ftp://ftp.cs.utexas.edu/pub/mooney/accel}\enspace.}
\cite{Ng1992kr,Ng1992thesis}
contains a knowledge base with 190 axioms of form \eqref{eqAx},
defines a set of sort names that observe the UNA,
and contains 50 instances (i.e., goals)
with between 5 and 26 atoms in a goal (12.6 atoms on average).
Axioms contain a single head and bodies vary between 1 and 11 atoms (2.2 on average).
\accel\ axioms contain no weights
and goal atoms contain no initial costs.
For experiments with \waobj\ we follow existing practice (cf.~\cite{Inoue2013})
and set initial costs to $ic(o) \eqs 100\$$ for all $o \ins O$
and for each axiom we set weights to sum up to $1.2$,
i.e., we set $w_i \eqs 1.2 / r$, $1 \les i \les r$.

In addition to axioms, goals, and sort names,
\accel\ contains constraints that forbid certain combinations of atoms to become abduced
at the same time (assumption nogoods)
and constraints that enforce functionality
for certain predicate symbols (unique slot axioms).
We next give two examples.
\begin{example}
  An example for an assumption nogood is,
  that we are not allowed to abduce an event $G$
  to be part of a `go' event $S$,
  and at the same time abduce that a person $P$ is the `goer' of $G$.
  \begin{align}
    \label{eqExAssumption}
    \not\exists S, G, P: \{ \mi{go\_step}(S,G), \mi{goer}(G,P) \} \ins H \text{ for all $H \ins \hcH$}
  \end{align}
  An example for a unique slot axiom is,
  that the `goer' of an event must be unique.
  \begin{align}
    \label{eqExUniqueSlot}
    \not\exists G, P_1, P_2: P_1 \neq P_2 \lands \{ \mi{goer}(G,P_1), \mi{goer}(G,P_2) \} \ins H \text{ for all $H \ins \hcH$}
  \end{align}
  \hfill$\square$
\end{example}

\subsection{Answer Set Programming}

We assume familiarity with ASP
\cite{Gelfond1988,Lifschitz2008,Eiter2009aspprimer,Gebser2012aspbook}
and give only brief preliminaries of \hex\ programs \cite{Eiter2015hex}
which extend the ASP-Core-2 standard \cite{Calimeri2012}.
We will use programs with (uninterpreted) function symbols,
aggregates, choices, and weak constraints.

\leanparagraph{Syntax.}
Let $\cC$, $\cX$, and $\cG$ be mutually disjoint sets
of \emph{constants}, \emph{variables}, and \emph{external predicate} names,
which we denote with first letter in \rev{upper case, lower case}{lower case, upper case},
and starting with \quo{\,\&\,}, respectively.
Constant names serve as constant terms, predicate names, and names for uninterpreted functions.
The set of \emph{terms} $\cT$ is recursively defined,
it is the smallest set containing $\bbN \cups \cC \cups \cX$
as well as uninterpreted function terms of form $f(t_1,\ldots,t_n)$ where
$f \ins \cC$ and $t_1,\ldots,t_n \ins \cT$.
An {\em \rev{}{ordinary }atom} is of the form $p(t_1,\dots,t_n)$,
where $p \ins \cC$, $t_1,\dots, t_n \ins \cT$, and
$n\geq 0$ is the \emph{arity} of the atom.
An {\em aggregate atom} is of the form $X \eqs \#\mi{agg} \{\ e_1 ; \ldots ; e_k \ \}$
with variable $X \ins \cX$, aggregation function $\#\mi{agg} \ins \{ \#\mi{min}, \#\mi{max} \}$,
$k \ges 1$ and each aggregate element $e_i$, $1 \les i \les k$,
is of the form $t \cols a$ or $t$ with $t \ins \cT$ and $a$ an atom.
An {\em external atom}\/ is of the form
$\amp{f}[\rev{Y_1,\ldots,Y_n}{y_1,\dots,y_n}](\rev{X_1,\dots,X_m}{x_1,\dots,x_m})$,
where $\rev{Y_1,\ldots,Y_n,X_1,\dots,X_m}{y_1,\dots,y_n,x_1,\dots,x_m} \ins \cT$ are two lists of terms
(called {\em input} and {\em output} lists, resp.),
and $\amp{f} \in \cG$ is an external predicate name.
An external atom provides a way for deciding the
truth value of an output tuple depending on the input tuple and a
given interpretation.
A term or atom is {\em ground} if it contains no sub-terms that are variables.

A  {\em rule $r$} is of the form
$ \alpha_1\lor\cdots\lor\alpha_k
  \leftarrow \beta_1, \dots, \beta_n,
  \naf\, \beta_{n+1},\dots,\naf\,\beta_{m}$
where $m,k\geq 0$, $\alpha_i$, $0 \les i \les k$ is an \rev{}{ordinary }atom
and $\beta_j$, $0 \les j \les m$ is an atom, \rev{aggregate atom, or external atom.}%
{and we let $H(r) = \{ \alpha_1,\ldots,\alpha_k\}$
and $B(r) = \{ \beta_1, \dots, \beta_n,
  \naf\,\beta_{n\pluss 1},\dots,\naf\,\beta_m \}$.}
A {\em program} is a finite set $P$ of rules.
A rule $r$ is a {\em constraint}, if $k \eqs 0$ %
and $m \neqs 0$, %
\rev{}{and }a {\em fact} if $m \eqs 0$%
\rev{,
and  {\em nondisjunctive}, if $k \les 1$. %
We call $r$ {\em ordinary}, if it contains only ordinary atoms.
We call a program $P$ \emph{ordinary} (resp., \emph{nondisjunctive}),
if all its rules are ordinary (resp., nondisjunctive)}{}.

A {\em weak constraint} is of form
${\wlar}\, \beta_1, \dots, \beta_n, \naf\, \beta_{n+1},\dots,
\naf\,\beta_{m}.\ [w@1,t_1,\ldots,t_k]$
where all $\beta_j$ are atoms\rev{, aggregate atoms, or external atoms}{},
and all $t_i$ are terms such that each variable in some $t_i$ is contained in some $\beta_j$
(note that $1$ in $w@1$ shows the `level' which we do not use).

\leanparagraph{Semantics.}
Semantics of a \hex\ program $P$ are defined
using its Herbrand Base $\HBP$ and its
ground instantiation $\grnd(P)$.
An aggregate literal in the body of a rule
accumulates truth values from a set of atoms,
e.g., $C \eqs \#min \{ 4 ; 2 : p(2) \}$
is true wrt.\ an interpretation $I \subseteqs \HBP$
iff $p(2) \ins I$ and $C \eqs 2$ or $p(2) \notins I$ and $C \eqs 4$.
Using the usual notion of satisfying a rule given an interpretation,
the FLP-reduct~\cite{Faber2011} $\fP^I$
reduces a program $P$ using an answer set candidate $I$:
$\fP^I \eqs \{ r \ins \grnd(P) \mids I \modelss B(r) \}$.
$I$ is an answer set of $P$ ($I \ins \AS(P)$) iff
$I$ is a minimal model of $\fP^I$.
Weak constraints define that an answer set $I$
has cost equivalent to the term $w$ for each distinct tuple $t_1,\ldots,t_k$
of constraints that have a satisfied body wrt.\ $I$.
Answer sets of the lowest cost are preferred.

\leanparagraph{Safety and Splitting.}
Programs must obey \emph{syntactic safety restrictions}
(see \cite{Calimeri2012})
to ensure a finite instantiation.
In presence of loops over external atoms,
\hex\ programs additionally must obey restrictions
to ensure finite instantiation.
A \emph{splitting set} \cite{Lifschitz1994} of a program $P$ is any set $U$ of literals such that,
for every rule $r \ins P$, if $H(\rev{P}{r}) \caps U \neqs \emptyset$ then $B(\rev{P}{r}) \subseteqs U$.
The set of rules $r \ins P$ such that $B(\rev{P}{r}) \subseteqs U$ is called the bottom $b_U(P)$ of $P$ relative to $U$.
Given splitting set $U$ of program $P$,
$I \ins \AS(P)$ iff $I \eqs X \cups Y$
where $X \caps Y \eqs \emptyset$,
$X \ins \AS(b_U(P))$,
and $Y \ins \AS(e_U(P \setminus b_U(P), X))$
where $e_U(Q,J)$ partially evaluates $Q$ wrt.\ $J$.
Splitting sets were lifted to \hex\ in
\cite{Eiter2006hex,Eiter2015hex}.

\leanparagraph{Syntactic Sugar.}
Anonymous variables of form \quo{$\_$}
are replaced by new variable symbols.
Choice constructions can occur instead of rule heads,
they generate a set of candidate solutions if the rule body is satisfied;
e.g., $1 \les \{ p(a) ; p(b) ; p(c) \} \les 2$ in the rule head
generates all solution candidates where at least 1 and at most 2 atoms of the set are true.
The bounds can be omitted.
In choices, the colon symbol \quo{:} can be used to generate a list of elements (similar as in aggregates),
for example $\{ p(X) : q(X), \naf\ r(X) \}$
encodes a guess over all $p(X)$ such that $q(X)$ is true and $r(X)$ is not true.

\section{ASP Encodings}
\label{secMainEncodings}

We next describe \rev{our }{ASP} encodings for modeling abduction
with partial UNA and value invention\rev{ in ASP}{}.
\rev{}{All encodings consist of a deterministic part
that instantiates all atoms that can potentially used
to build a hypothesis, i.e., to justify the goal.
Some encodings also explicitly represent inferences
leading to these atoms.
All encodings guess an equivalence relation over
all terms in these \quo{potentially interesting} atoms,
to handle term equivalence.
In \bwd\ encodings,
the actually used proof graph is nondeterministically guessed,
while encoding \fwda\ performs a guess of abduced atoms
and checks if the goals become true given these atoms.
Variations of \bwd\ encodings perform factoring
in different ways.
We next give common aspects of all encodings,
then give each encoding in detail,
and finally provide a summary of differences between encodings (Table~\ref{tblEncodings}).

A detailed example containing a rewriting
of our running example and an answer set corresponding to Figure~\ref{figExWA}
is given in the appendix.
}

We represent an atom of the form $p(a,b)$ as a term $c(p,a,b)$%
\rev{and}{, which allows us to quantify over predicates using ASP variables
(e.g., $c(P,X,Y)$).
We} represent each atom in a goal \eqref{eqObs}
as a fact
\begin{align}
  \label{eqGoalFact}
  \migoal(c(o_1,v_1,\ldots,v_k)).
\end{align}
where $v_i \notins S$ are constants corresponding to
existentially quantified variables $V_i$.

We mark each sort name $s \ins S$ using a fact
\begin{align}
  \label{eqSortFact}
  \misortname(s).
\end{align}

\begin{example}[continued]
  Goal and sort names of our running example
  are represented as
  \begin{align*}
          &\migoal(c(\miname,m,\mimary)).
    \quad \migoal(c(\milost,m,f)).
    \quad \migoal(c(\mifatherof,f,m)). \\
          &\migoal(c(\miinst,s,\mifemale)).
    \quad \migoal(c(\miis,s,\midepressed)).
    \quad \misortname(\miperson). \\
          &\misortname(\mimale).
    \quad \misortname(\mifemale).
    \quad \misortname(\midead).
    \quad \misortname(\midepressed).
  \end{align*}
  \hfill$\square$
\end{example}

\subsection{Rules common to all encodings}
\label{secCommonRules}

Goals are potential interesting facts%
\rev{}{, i.e., potential nodes of the proof graph}:
\begin{align}
  \label{eqPotential}
  \mipot(X) \lars \migoal(X).
\end{align}
Potential interesting facts provide
potentially relevant terms in the Herbrand Universe (HU).
\begin{align}
  \label{eqHu1}
  hu(X) &\lars \mipot(c(\_,X,\_)). \\
  \label{eqHu2}
  hu(X) &\lars \mipot(c(\_,\_,X)).
\end{align}
Note that \eqref{eqHu1} and \eqref{eqHu2} assume,
that all atoms in $B$ and $O$ have arity 2.
This assumption is made only in these two rules,
which can be generalized easily to arbitrary arities.

We call terms in HU that are not sort names \quo{User HU},
represent them in predicate $\miuhu$,
and guess a relation $\mieq$ among pairs of these terms.
\begin{align}
  \label{eqUhu}
  \miuhu(X) &\lars hu(X), \naf\ \misortname(X). \\
  \label{eqEqGuess}
  \{\ eq(A,B) : \miuhu(A),\ \miuhu(B),\ A \neqs B\ \} &\lars.
\end{align}
Relation $\mieq$ holds symmetric on HU,
and it is a reflexive, symmetric, and transitive
(equivalence) relation.
\begin{align}
  \label{eqEqReflexive}
  eq(A,A) &\lars hu(A). \\
  \label{eqEqSymmetric}
  &\lars eq(A,B), \naf\ eq(B,A). \\
  \label{eqEqTransitive}
  &\lars eq(A,B), eq(B,C), A \rev{\lts}{\neqs} B, B \rev{\lts}{\neqs} C, \rev{}{A \neqs C,} \naf\ eq(A,C).
\end{align}
\rev{}{Note, that we will later create instantiations of
constraint~\eqref{eqEqTransitive} in a lazy manner (on-demand),
therefore we use a constraint to ensure transitivity,
and not a rule.}

\subsection{\bwd: Defining Back-chaining Proof Graph, Guess Active Parts}
\label{secBwd}

We next encode the maximal back-chaining proof graph
from Definition~\ref{defHypGraph} in ASP by:%
\bipe{(i)}
\item
  deterministically defining the maximum possible
  potential proof graph by back-chaining from the goal
  and creating new constants when required;
\item
  guessing which parts of the proof graph are used,
  i.e., which atoms are back-chained over which axioms,
  and which bodies of axioms must therefore be justified;
\item
  factor atoms with other atoms
  and mark the remaining atoms as abduced.
\eipe

\paragraph{Potential Proof Graph.}
Building the potential proof graph
is realized by rewriting each axiom of form \eqref{eqAx}
into a deterministic definition of potential inferences
from the axiom's head atom,
and defining which body atoms become part of the
hypothesis due to such an inference.
\smallskip

\rev{}{We first give this rewriting as an example and then formally.}
\begin{example}
\label{exUFSkolemizationRewriting}
\rev{For example a}{A}xiom \eqref{eqEx4} \rev{}{from our running example }is translated into the following rules.
\begin{eqnarray}
  &&\mimayinfervia(r_1,c(\miis,X,\midepressed),l(Y)) \lars \notag \\
  \label{eqEx4rewriteHead}
  &&\hspace*{16em} \mipot(c(\miis,X,\midepressed)), Y \eqs \miskolem(r_1,``Y",X). \\
  &&\miinferenceneeds(c(\miis,X,\midepressed),r_1,c(\miimportantfor,Y,X)) \lars \notag\\
  \label{eqEx4rewriteBody1}
  &&\hspace*{16em} \mimayinfervia(r_1,c(\miis,X,\midepressed),l(Y)). \\
  &&\miinferenceneeds(c(\miis,X,\midepressed),r_1,c(\miis,Y,\midead)) \lars \notag \\
  \label{eqEx4rewriteBody2}
  &&\hspace*{16em} \mimayinfervia(r_1,c(\miis,X,\midepressed),l(Y)).\quad
\end{eqnarray}
Here $r_1$ is a unique identifier for axiom
\eqref{eqEx4}.
Rule \eqref{eqEx4rewriteHead} defines all possible substitutions of the axiom,
including value invention via Skolemization
which is represented in the last argument of
$\mimayinfervia$ in the term $l(\cdots)$.
Rules \eqref{eqEx4rewriteBody1} and \eqref{eqEx4rewriteBody2}
define which body atoms become part of the hypothesis
because of back-chaining.
Note that Skolemization is here realized with an
uninterpreted function term $\miskolem(r_1,``Y",X)$
that takes as arguments the unique axiom identifier $r_1$,
the name of the skolemized existential variable $``Y"$
(to skolemize several variables in one axiom independently),
and all variables (here only $X$) in the body of the axiom.
\hfill$\square$
\end{example}

For each body atom \rev{$\mi{Body}$ }{}that can be added
to the hypothesis by an inference,
$\miinferenceneeds(\mi{Head}$, $\mi{Rule},\mi{Body})$
is defined.
To allow back-chaining from $\mi{Body}$,
we define $\mi{Body}$ as potentially interesting.
\begin{eqnarray}
  \label{eqNeedPotential}
  &\mipot(P) \lars \miinferenceneeds(\_,\_,P)
\end{eqnarray}
Axioms rewritten as in \eqref{eqEx4rewriteHead}--\eqref{eqEx4rewriteBody2}
together with \eqref{eqPotential} and \eqref{eqNeedPotential}
form a deterministic ASP program,
that, given a set of goal atoms $\migoal(A)$,
defines the union of all possible proof graphs.

This graph is finite under the assumption that the knowledge base
is acyclic, i.e., that no circular inferences are possible
over all axioms in the knowledge base.
(For reasons of presentation we will maintain this assumption
while presenting the basic encodings
and eliminate the assumption in Section~\ref{secFlexibleValueInvention}.)

For readability we gave the rewriting
\rev{as an example}{%
in Example~\ref{exUFSkolemizationRewriting}}.
Formally, an axiom of form \eqref{eqAx} is rewritten
into
\begin{eqnarray}
  &&\mimayinfervia(a,c(q,Y_1,\ldots,Y_m),l(Z_1,\ldots,Z_v)) \lars
      \notag \\
  &&\rulebodyhspace
      \mipot(c(q,Y_1,\ldots,Y_m)),\
      Z_1 \eqs s(a,1,Y_1,\ldots,Y_m),\
      \ldots, Z_v \eqs s(a,v,Y_1,\ldots,Y_m)\quad \notag \\
  &&\miinferenceneeds(c(q,Y_1,\ldots,Y_m),a,c(p_i,X^i_1,\ldots,X^i_{k_i})) \lars \label{eqFormalBwd} \\
  &&\rulebodyhspace \mimayinfervia(a,c(q,Y_1,\ldots,Y_m),l(Z_1,\ldots,Z_v))
  \qquad\qquad \text{for } i \ins \{1,\ldots,r\} \notag
\end{eqnarray}
where $a$ is a unique identifier for that particular axiom,
$Z_1,\ldots,Z_v \eqs \cX \setminuss \cY$
is the set of variables occurring in the body but not in the head,
and the second argument of the uninterpreted function $s(\cdot)$
is a unique identifier for each skolemized variable in this axiom.

\begin{lemma}
  \label{thmPbptProofGraph}
  Given an abduction instance $A \eqs (B,O,S)$,
  let $P_\mi{bos}(A) \eqs P_b \cup P_o \cup P_s$
  where $P_b$ is the rewriting of each axiom in $B$ as \eqref{eqFormalBwd},
  $P_o$ the rewriting of $O$ as \eqref{eqGoalFact},
  and $P_s$ the rewriting of $S$ as \eqref{eqSortFact}.
  Let $P_\mi{bpt}(A) = P_\mi{bos}(A) \cups \{$\eqref{eqPotential}, \eqref{eqNeedPotential}$\}$.
  Then $P_\mi{bpt}(A)$ has a single answer set $I$
  that represents the union of all proof graphs
  of all hypotheses of $A$
  that are generated by back-chaining according to
  Def.~\ref{defHypGraph},
  with nodes $\{ P \mids \mipot(P) \ins I \}$
  and edges
  $\{ (Q,P) \mids \mimayinfervia(R,P,L) \ins I,
      \miinferenceneeds(P,R,Q) \ins I \}$.
\end{lemma}
\newcommand{\myProofThmPbptProofGraph}{
  $P_\mi{bpt}(A)$ does not contain $\naf$
  and its rules are not expanding term depth
  except for the first rule of \eqref{eqFormalBwd}.
  As $B$ is acyclic,
  by construction of $P_\mi{bpt}(A)$
  there are no loops over \eqref{eqFormalBwd},
  $\grnd(P_\mi{bpt}(A))$ is finite,
  and $\AS(P_\mi{bpt}(A)) = \{ I \}$
  has a single answer set $I$.
  The representation of proof graphs is achieved
  as follows:
  atoms $A \eqs p(a,b)$ that can be back-chained
  are represented as $\mipot(c(p,a,b)) \ins I$:
  We proceed by induction on the distance $d$ of back-chaining from observations.
  (Base: $d \eqs 0$)
  Due to \eqref{eqPotential} all atoms $p(a,b) \ins O$
  are true as $\mipot(c(p,a,b)) \ins I$.
  (Step: $d \Rightarrow d\pluss 1$)
  Assuming that $A = q(a,b)$, $A \ins H$, $H \ins \hcH$,
  is represented as $\mipot(c(q,a,b)) \ins I$,
  this causes an instantiation of the first rule
  in \eqref{eqFormalBwd},
  which defines $\mimayinfervia(r,c(q,a,b),l(z_1,\ldots,z_v))$ true in $I$.
  This represents potential backchaining from $q(a,b)$
  over an axiom identified by $r$
  using substitution $\theta = \{ Z_1 \mapsto z_1, \ldots, Z_v \mapsto z_v \}$
  where $Z_1,\ldots,Z_v$ are variables occurring only in the body of $r$.
  Truth of $\mimayinfervia(r,c(q,a,b),l(z_1,\ldots,z_v))$
  causes truth of $\miinferenceneeds(c(q,a,b),r,c(p_1,x_1^1,x_2^1)$ where $p_i$ is the predicate and $x_i^j$ are the substituted variables at position $i$ in body atom $j$ of axiom $r$, analogous to backward chaining in Def.~\ref{defHypotheses}.
  Due to truth of $\miinferenceneeds(c(q,a,b),r,c(p_i,x_i^j,x_{i+1}^j))$ and due to \eqref{eqNeedPotential},
  all body atoms $\theta(p_i(X^i_1,\ldots,X_{k_i}^i))$
  added to some $H \ins \hcH$ in Def.~\ref{defHypotheses}
  become represented as $\mipot(c(p_i,x_i^j,x_{i+1}^j) \ins I$.
  (Conclusion)
  We conclude that $I$ contains all atoms $p(a,b)$
  in hypotheses generated from observations
  represented as $\mipot(c(p,a,b)) \ins I$.
  Moreover, $\mimayinfervia(\cdots)$
  represents potential back-chaining
  and $\miinferenceneeds(\cdots)$ represents
  the body atoms that are added to a hypothesis
  by a particular backchaining.
}

\paragraph{Representing a Hypothesis.}
Based on the potential proof graph defined above,
we now formulate in ASP the problem of
guessing a hypothesis,
i.e., selecting a connected part of the potential proof graph
as a solution to the abduction problem.
Nodes of the proof graph are represented as $\mitrue(\cdot)$.

If an atom $P$ is a goal, it is true.
\begin{align}
  \label{eqGoalTrue}
  \mitrue(P) \lars \migoal(P).
\end{align}
If an atom $P$ is true,
it is back-chained ($\miinfer(\cdot)$),
factored, or abduced (the latter two represented as $\mifai$).
\begin{align}
  \label{eqTrueInferFai}
  1 \les \{\ \miinfer(P)\ ;\ \mifai(P)\ \} \les 1 \lars \mitrue(P).
\end{align}
\rev{If an atom $P$ is inferred, it is }%
{Each atom $P$ that is marked as inferred in the proof graph,
has to be} back-chained via exactly \rev{an }{one }axiom $R$.
\begin{align}
  \label{eqInferViaWhat}
  1 \les \{\ \miinfervia(R,P) : \mimayinfervia(R,P,\_)\ \} \les 1 \lars \miinfer(P).
\end{align}
If back-chaining an atom $P$ would add body atom $Q$ to the hypothesis,
then we define $Q$ as true.
\begin{align}
  \label{eqNeedInferTrue}
  \mitrue(Q) \lars \miinfervia(R,P), \miinferenceneeds(P,R,Q).
\end{align}

This encoding guesses the back-chaining part
of a particular proof graph and hypothesis.

\begin{proposition}
  \label{thmBwdGuessProofGraph}
  Given an abduction instance $A = (B,O,S)$,
  let $P_\mi{gp}$ consist of rules
  \eqref{eqGoalTrue}--\eqref{eqNeedInferTrue}.
  Then answer sets $\AS(P_\mi{bpt}(A) \cups P_\mi{gp})$
  correspond 1-1 with proof graphs $G$ induced
  by hypotheses $H \ins \hcH(A)$ via back-chaining:
  edges $E(G)$ are represented as $\miinfervia(\cdot,\cdot)$,
  nodes $N(G)$ as $\mitrue(\cdot)$,
  back-chained atoms as $\miinfer(\cdot)$,
  and other atoms in $\mifai(\cdot)$.
\end{proposition}
\newcommand{\myProofThmBwdGuessProofGraph}{
  We write $P_\mi{bpt}$ for $P_\mi{bpt}(A)$.
  $P_\mi{bpt}$ is a bottom of $P_\mi{bpt} \cups P_\mi{gp}$
  and therefore each $I \ins \AS(P_\mi{bpt} \cups P_\mi{gp})$
  is such that $I = I' \cups I_g$ where $I' \ins \AS(P_\mi{bpt})$
  and $I_g$ contains only predicates $\miinfer$, $\mifai$, $\miinfervia$, $\mitrue$.
  \eqref{eqGoalTrue} defines for all $o \ins O$,
  $o = p(a,b)$, that $\mitrue(c(p,a,b)) \ins I_g$.
  \eqref{eqTrueInferFai} defines that every atom $P$
  with $\mitrue(P) \ins I_g$,
  either $\miinfer(P) \ins I_g$ or $\mifai(P) \ins I_g$
  (i.e., two answer set candidates are generated).
  \eqref{eqInferViaWhat} defines that every atom $P$
  with $\miinfer(P) \ins I_g$ is marked as inferred
  via a particular axiom along an edge
  $\mimayinfervia(R,P,Z) \ins I'$
  of the potential proof graph,
  and represents this inference as $\miinfervia(R,P)$
  where $R$ is the axiom identifier.
  \eqref{eqNeedInferTrue} defines that for each $\miinfervia(R,P)$
  the corresponding required body atoms $X^i_1,\ldots,X_{k_i}^i$
  represented as $\miinferenceneeds(P,R,c(p_i,X_1^i,\ldots,X_k^i) \ins I'$
  are defined as $\mitrue(c(p_i,X_1^i,\ldots,X_k^i)) \ins I_g$.
  As defining these as true, they again must be represented
  as $\miinfer$ or $\mifai$ due to \eqref{eqTrueInferFai}.
  Due to minimality of answer sets,
  all atoms marked as $\mitrue$, $\miinfer$, and $\mifai$
  are reachable from objectives $o \ins O$
  which are also represented as $\mitrue(\cdot) \ins I_g$.
  The set of hypotheses is inductively built from
  observations and any combination of backchaining
  over axioms from these observations and atoms obtained
  from backchaining,
  exactly as the set of atoms marked as $\mitrue$ in $I_g$
  is inductively defined from observations
  and any choice about back-chaining
  in \eqref{eqTrueInferFai}.
  Therefore, answer sets and proof graphs are in 1-1 correspondence.
}

\subsection{Factoring}
\label{secFactoring}

So far we merely encoded the back-chaining part of proof graphs.
It remains to deal with unification,
which allows us to identify those atoms in a hypothesis
that incur a cost
in \cardobj\ and \waobj\ because they must be abduced.
For that we use rules \eqref{eqHu1}--\eqref{eqEqTransitive}
which guess an equivalence relation $\mieq(\cdot,\cdot)$
over the Herbrand Universe such that
constants that are not sort names
can be equivalent with other constants.
\begin{lemma}
  \label{thmBwdGuessEq}
  Given an abduction instance $A = (B,O,S)$,
  let $P_\mi{eq}$ consist of rules
  \eqref{eqHu1}--\eqref{eqEqTransitive}.
  Then $\AS(P_{bpt}(A) \cups P_\mi{eq})$
  contains one answer set
  for each equivalence relation
  on the HU of $I \ins \AS(P_{bpt}(A))$
  represented in predicate $\mieq$
  such that sort names are singleton
  equivalence classes.
\end{lemma}
\newcommand{\myProofThmBwdGuessEq}{
  We write $P_\mi{bpt}$ for $P_\mi{bpt}(A)$.
  $P_\mi{eq}$ does not define predicates present in $P_{bpt}$,
  hence $P_{bpt}$ is a bottom wrt.\ $P_{bpt} \cups P_\mi{eq}$
  (see \cite{Lifschitz1994}) and
  $I' \ins \AS(P_{bpt} \cups P_\mi{eq})$
  is such that $I' = I \cups I_e$
  where $I \ins \AS(P_{bpt})$
  and $I_e$ contains only predicates
  $\mieq$, $\mihu$, and $\miuhu$.
  All constants $c$ in argument positions of hypotheses
  are represented in $\mihu(c)$ due to \eqref{eqHu1}--\eqref{eqHu2},
  those that are not sort names are additionally represented in
  $\miuhu(c)$ due to \eqref{eqUhu}.
  \eqref{eqEqGuess} guesses
  a relation $\mieq(c,c')$ a solution candidate
  for all pairs $(c,c')$ of constants
  with $c \neq c'$ that do not contain sort names.
  Finally, \eqref{eqEqReflexive}
  defines $\mieq$ to be reflexive for all constants
  (including sort names),
  and \eqref{eqEqSymmetric}/\eqref{eqEqTransitive}
  exclude answer sets where the represented relation
  $\mieq$ is not symmetric/transitive.
  Therefore, only those relations remain that are
  reflexive, symmetric, and transitive,
  i.e., equivalence relations.
}

Atoms that are not back-chained
are represented as $\mifai(P)$.
These must either be unified or abduced.

\subsubsection{\bwdg: Guess Factoring}

This method guesses whether an atom is factored or abduced
and represents for factored atoms
with which other atom they have been unified.
As the deterministically defined proof graph does not contain
factoring between inference steps,
we require factoring with inferred atoms
to obtain all possible proof graphs.
(We discuss and relax this restriction in Section~\ref{secBwda}.)

For an atom in $H$ that is not inferred,
we guess if it is factored or abduced.
\begin{align}
  \label{eqFactorOrAbduce}
  1 \les \{\ \mifactor(P)\ ;\ \miabduce(P)\ \} \les 1 \lars \mifai(P).
\end{align}
If a factored atom $A_1 \eqs c(P,S_1,O_1)$
unifies with an inferred atom $A_2 \eqs c(P,S_2,O_2)$
that is not below $A_1$ in the proof graph,
then represent that $A_1$ is factored via $A_2$.
\begin{eqnarray}
  &&\mifactorvia(c(P,S_1,O_1),c(P,S_2,O_2)) \lars
    \mifactor(c(P,S_1,O_1)), \miinfer(c(P,S_2,O_2)), \notag\\
  \label{eqGuessFactorInfer}
  &&\rulebodyhspace
    \naf\ \mibelow(c(P,S_2,O_2),c(P,S_1,O_1)), eq(S_1,S_2), eq(O_1,O_2).
\end{eqnarray}
We define $\mibelow(A_1,A_2)$ as a partial order over atoms
such that $A_1$ is below $A_2$
whenever inference of $A_1$ requires $A_2$,
and whenever $A_1$ is factored via $A_2$.
Intuitively, \quo{below} can be read as
\quo{closer to goal nodes}.
\begin{eqnarray}
  \label{eqBelow1}
  &&\mibelow(P,Q) \lars \miinfervia(R,P), \miinferenceneeds(P,R,Q). \\
  \label{eqBelow2}
  &&\mibelow(A,C) \lars \mibelow(A,B), \mibelow(B,C). \\
  \label{eqBelow3}
  &&\mibelow(P,Q) \lars \mifactorvia(P,Q).
\end{eqnarray}
Note that without $\naf\ \mibelow(\cdots)$
in \eqref{eqGuessFactorInfer},
we would obtain cyclic proof graphs
where an atom justifies itself.
This would affect all objective functions
we study, therefore we need to eliminate such cases.

If a factored atom unifies with an abduced atom,
represent that this is the case.
\begin{eqnarray}
  &&\mifactorvia(c(P,S_1,O_1),c(P,S_2,O_2)) \lars \notag \\
  &&\rulebodyhspace
    \mifactor(c(P,S_1,O_1)), \miabduce(c(P,S_2,O_2)),
    eq(S_1,S_2), eq(O_1,O_2).
\end{eqnarray}
Finally, we require that all factored atoms
are unified with another atom.
\begin{align}
  \label{eqFactorOk}
  &&\mifactorOk(P) \lars \mifactorvia(P,\_). \\
  &&\lar\, \mifactor(P), \naf\ \mifactorOk(P).
\end{align}
We do not prove correctness of \bwdg,
as encoding \bwda\ is similar and has better performance.

\subsubsection{\bwdai: Abduced/Inferred Cluster Factoring}

As an alternative to guessing
which atoms are factored and which are abduced,
we next define deterministically,
that every atom that can be factored with an inferred atom
must be factored with that atom,
and that all remaining sets $X$ of atoms
that unified wrt.\ $\mieq$
are factored with the (lexicographically) smallest
atom in that equivalence class $X$ of atoms,
in the following called \quo{cluster}.

To that end, instead of \eqref{eqGuessFactorInfer}
we use the following rule.
\begin{eqnarray}
  &&\mifactorviai(c(P,S_1,O_1),c(P,S_2,O_2)) \lars
    \mifai(c(P,S_1,O_1)), \miinfer(c(P,S_2,O_2)), \notag\\
  \label{eqIAClusterFactorInfer}
  &&\rulebodyhspace
    \naf\ \mibelow(c(P,S_2,O_2),c(P,S_1,O_1)), eq(S_1,S_2), eq(O_1,O_2).
\end{eqnarray}
We represent atoms that are factored via inferred atoms
using predicate $\mifactori$
and we represent what remains to be factored or abduced in $\mifa$.
Moreover, we define that $\mifactorviai$ entails $\mifactorvia$.
\begin{align}
  \mifactori(P) &\lars \mifactorviai(P,\_). \\
  \mifa(P) &\lars \mifai(P), \naf\ \mifactori(P). \\
  \mifactorvia(A,B) &\lars \mifactorviai(A,B).
\end{align}
Next we deal with these remaining atoms:
we define a partial order over atoms that unify under equivalence $\mieq$
and factor these clusters with the (lexicographically) smallest element as follows.
\begin{align}
  \mifactorcluster(c(P,S_2,O_2),c(P,S_1,O_1)) \lars& \mifa(c(P,S_1,O_1)), \mifa(c(P,S_2,O_2)), eq(S_1,S_2), \notag \\
    & c(P,S_1,O_1) \lts c(P,S_2,O_2), eq(O_1,O_2). \label{eqCluster} \\
  \mifactorclusterabove(A) \lars&
    \mifactorcluster(A,\_). \label{eqClusterNonlargest} \\
  \mifactorvia(A,B) \lars& \mifactorcluster(A,B), \notag \\
    & \naf\ \mifactorclusterabove(B). \label{eqClusterFactorVia}
\end{align}
Note that \eqref{eqCluster} defines the partial order,
\eqref{eqClusterNonlargest} represents elements that are not the smallest in the cluster,
and \eqref{eqClusterFactorVia} maps the partial order into
$\mifactorvia$ using the smallest element as the atom
that all others are unified with.
Finally, we define $\mibelow$
using rules \eqref{eqBelow1}--\eqref{eqBelow3},
and we define that every hypothesis atom that could
not be factored is abduced.
\begin{align}
  \mifactor(P) \lars& \mifactorvia(P,\_). \label{eqClusterFactor} \\
  \miabduce(P) \lars& \mifa(P), \naf\ \mifactor(P). \label{eqClusterAbduce}
\end{align}

Note that this encoding represents a restricted set of solutions
compared to the previous one,
however because of the symmetry of unification,
by fixing the direction of factoring
we merely canonicalize solutions and cannot lose optimal solutions.
\begin{example}[continued]
  \rev{For reversing factoring arcs between
  an abduced and a non-abduced atom,}%
  {To illustrate, that the factoring method of \bwdai\
  does not eliminate optimal solutions, }%
  consider the arc between \rev{}{the abduced atom}
  $\mifatherof(f,m)$ and
  \rev{}{the factored atom}
  $\mifatherof(f,n_2)$ in Figure~\ref{figExWA}:
  \rev{this makes the former factored and the latter abduced}%
  {if we reverse this arc, then the former becomes factored
  and the latter abduced.
  The number of abduced atoms stays the same,}
  therefore \cardobj\ is not affected;
  reachability stays the same so \accelobj\ is not affected;
  and costs propagate the other direction:
  \rev{}{$\mifatherof(f,m)^{\mb{48}\text{\$},\,\underline{100\text{\$}}}$
  obtains cost via \eqref{eqEx5} and initial goal cost;
  $\mifatherof(f,n_2)^{\mi{48}\textit{\$},\,\mb{57}\textbf{\$}}$ obtains cost from factoring and via \eqref{eqEx1},}
  so \waobj\ \rev{is not affected}{remains unchanged}.

  Similarly, the factoring edge between
  $\miname(m,\mimary)$ and $\miname(s,\mimary)$
  could be reversed:
  then
  \rev{$\micost(\miname(m,\mimary)) \eqs \{ 100\$ \}$
  (because it is a goal),}{
  $\miname(m,\mimary)^{\underline{100\$}}$
  obtains only initial goal cost}
  and we would abduce
  $\miname(s,\mimary)\rev{}{^{\mi{100}\textit{\$},\,\mb{120}\textbf{\$}}}$
  \rev{(it obtains $120\$$ from inference and $100\$$ from factoring)}{}
  and minimum cost of \waobj\ remains $100\$$
  \rev{}{for these nodes},
  moreover the number of abduced atoms (\cardobj)
  and reachability (\accelobj) stays the same.

  For reversing factoring arcs between two non-abduced atoms,
  consider reversing the arc between
  $\miimportantfor(f,m)$ and $\miimportantfor(n_1,m)$
  in Figure~\ref{figExWA}:
  reversing that arc makes the former atom factored
  \rev{}{(with costs $40\$$) }%
  and the latter abduced \rev{}{(with costs $40\$$ and $60\$$)},
  and back-chaining using \eqref{eqEx5}
  can be done from $\miimportantfor(n_1,m)$ instead,
  which yields the abduced atom $\mifatherof(n_1,m)$%
  \rev{ of same cost $48\$$}{$^{\mb{48}\textbf{\$},\,\mi{57}\textit{\$},\,\underline{100\$}}$}
  instead of $\mifatherof(f,m)$.
  Note that $\mifatherof(f,n_2)$ can still be factored
  with that new abduced atom as $f \eqs n_1$.
  \hfill$\square$.
\end{example}
As we are solving an optimization problem,
dealing with a subset of solutions that has been canonicalized
(by enforcing an arbitrary order on factoring)
can be an advantage for efficiency,
as certain symmetric solutions are automatically excluded.

We do not prove correctness of this factoring variant,
as the following variant has better performance.

\subsubsection{\bwda: Abduced Cluster Factoring}
\label{secBwda}

Finally, we apply an even stronger canonicalization to the proof graph:
we assume factoring only happens with abduced atoms.
We first show,
that every proof graph,
that contains factoring with an inferred atom,
can be transformed into a proof graph where inferences between factored atom and abduced atoms
is duplicated and all factoring is done with abduced atoms.

\begin{example}[continued]
  The proof graph in Figure~\ref{figExWA}
  can be transformed into a graph where factoring
  happens only with abduced atoms:
  instead of factoring atom
  $\miimportantfor(n_1,m)^{\mBcost{60}}$
  with atom $\miimportantfor(f,m)^{\mBcost{40}\rev{}{,\,\mIcost{60}}}$,
  we can back-chain from the former
  over axiom \eqref{eqEx5}
  which yields the atom $\mifatherof(n_1,m)^{\mBcost{72}}$
  in the graph.
  This atom can now be factored with
  $\mifatherof(f,m)$ at the top,
  which \rev{}{obtains the set $\{48\$,\,72\$,\,100\$\}$ of costs and therefore}
  keeps the same \rev{}{minimum }cost.
  \hfill$\square$
\end{example}

Importantly, the metrics we consider
do not change when we perform this canonicalization.

\begin{proposition}
  \label{thmFactoringAbducedSameCost}
  Given a proof graph $G$ of a hypothesis $H \ins \hcH(A)$ of an abduction instance $A$,
  there is a proof graph $G'$ and hypothesis $H'$
  of same cost wrt.\
  \cardobj, \accelobj, \waobj\ where factoring
  is only performed with atoms in $A(G')$.
\end{proposition}
\newcommand{\myProofThmFactoringAbducedSameCost}{
  We show how to push factoring edges closer to abduced atoms
  without changing the objective function value.
  As the graph is acyclic, this operation can be continued
  until we only unify with abduced atoms.

  Given an atom $P \ins H$ and an edge $(P,Q) \ins E(G)$
  of factoring $Q$ with $P$ using substitution $\theta$,
  and $P \notins A(G)$.
  Then either
  (i) $P$ is factored with $Q'$, i.e., $(Q',P) \ins E(G)$,
  or (ii) $P$ is back-chained on axiom $r$ with $k$ body atoms,
  i.e., $(P_i',P) \ins E(G)$ for $1 \les i \les k$.
  In case (i) we can factor $Q$ with $Q'$ instead of with $P$,
  which pushes factoring one edge closer to abduced atoms.
  This does not affect \cardobj\ as $A(G) \eqs A(G')$,
  reachability stays the same so \accelobj\ is not affected,
  and the minimal cost of $P$ and $Q$ is
  propagated to $Q'$ as before the change,
  so \waobj\ is not affected.
  In case (ii) we can back-chain from $Q$ with axiom $r$,
  creating edges $(\theta^{-1}(P_i'),Q)$,
  adding nodes $\theta^{-1}(P_i')$ to $G'$
  if they do not exist
  (implicitly they are already contained in $H$ due to equivalences)
  and adding factoring edges
  from $(P_i',\theta^{-1}(P_i'))$ to $G'$.
  This reduces the number of inference edges
  between factored and abduced atoms in the graph by 1.
  Similar as before,
  reachability and number of abduced atoms stays constant.
  For \waobj, cost might increase for
  $\theta^{-1}(P_i')$ but stays the same for $Q'$.
  Therefore, we do not lose optimal solutions
  by restricting factoring to abduced atoms.
}
\begin{proof}
  \myProofThmFactoringAbducedSameCost
\end{proof}

By a similar argument,
the order of factoring in such a proof graph does not matter,
so we can also canonicalize factoring direction.
\begin{proposition}
  \label{thmFactoringSwappable}
  Given a proof graph $G$ of a hypothesis $H$
  where factoring is only performed with atoms in $A(G)$,
  an abduced atom $P \ins A(G)$,
  and atoms $Q_1,\ldots,Q_k$ that are factored with $P$.
  Then we can swap an arbitrary $Q_i$, $1 \les i \les k$,
  with $P$, factor all other $Q_j$, $j \neqs i$ with $Q_i$,
  factor $P$ with $Q_i$,
  and all objective functions stay the same.
\end{proposition}
\newcommand{\myProofThmFactoringSwappable}{
  As all $Q_1,\ldots,Q_k,P$ unify,
  we can arbitrarily choose one of them as
  representative and factor all others with it.
  This does not increase the number of abduced atoms in \cardobj,
  this does not affect reachability in \accelobj,
  and costs of all atoms are propagated to the chosen representative,
  and the minimum cost in \waobj\ stays the same.
}
\begin{proof}
  \myProofThmFactoringSwappable
\end{proof}

To realize this canonicalization,
we use rules
\eqref{eqCluster}--\eqref{eqClusterAbduce}
and add the following rule,
such that every atom that is not back-chained
is factored with abduced atoms if possible,
otherwise abduced.
\begin{align}
  \mifa(P) \lars& \mifai(P). \label{eqFaFai}
\end{align}

Note that this encoding does not require any guesses
to determine what is factored and what is abduced,
moreover there is no need for the definition of $\mibelow$.

\begin{proposition}
  \label{thmBwda}
  Given an abduction instance $A = (B,O,S)$,
  let $P_\bwda(A) \eqs
    P_\mi{bpt}(A) \cups P_\mi{gp} \cups P_\mi{eq} \cup P_c$
  where $P_c = \{$\eqref{eqCluster}--\eqref{eqClusterAbduce},\eqref{eqFaFai}$\}$,
  then answer sets $\AS(P_\bwda(A))$ of $P_\bwda(A)$
  are in 1-1 correspondence with proof graphs $G$
  and hypotheses $H \ins \hcH(A)$
  where factoring is performed
  only with $A(G)$
  and only with lexicographically smaller atoms.
\end{proposition}%
\newcommand{\myProofThmBwda}{
  Let $P_\mi{gpeq}(A) = P_\mi{bpt}(A) \cups P_\mi{gp} \cups P_\mi{eq}$,
  then $P_\mi{bpt}(A) \cups P_\mi{gp}$ and $P_\mi{bpt}(A) \cups P_\mi{eq}$
  are bottoms wrt.\ $P_\mi{gpeq}(A)$,
  and $P_\mi{gp}$ and $P_\mi{eq}$ do not have common head atoms,
  hence both Prop.~\ref{thmBwdGuessProofGraph} and Prop.~\ref{thmBwdGuessEq}
  apply to answer sets $I \ins \AS(P_\mi{gpeq}(A))$,
  viz.\ each $I$ is in 1-1 correspondence with some proof graph $G$ and hypothesis $H \ins \hcH(A)$
  and moreover represents some equivalence relation over all constants of the proof graph in $I$,
  moreover all proof graphs originating in back-chaining are covered.
  Furthermore, $P_\mi{gpeq}(A)$ is a bottom wrt.\ $P_\bwda(A)$,
  therefore each answer set $I' \ins \AS(P_\bwda(A))$ is such that $I' = I_\mi{gpeq} \cups I$
  where $I_\mi{gpeq} \ins \AS(P_\mi{gpeq}(A))$
  and $I$ contains predicates defined by $P_c$ based on $I_\mi{gpeq}$.
  $P_c$ contains only stratified negation.
  \eqref{eqFaFai} defines $\mi{fa}(P)$ to be true iff $\mi{fai}(P) \ins I_\mi{gpeq}$,
  hence iff atom $P$ is not inferred in $H$.
  For all atoms $P,Q \ins H$ that are not inferred,
  \eqref{eqCluster} defines $\mifactorcluster(P,Q)$
  to be true if $P$ and $Q$ unify under $\mi{eq}$ represented in $I_\mi{gpeq}$,
  $Q$ is lexicographically smaller than $P$,
  and neither $P$ nor $Q$ were
  back-chained in $G$,
  i.e., they would be abduced unless they can be factored.
  Note that, given a set $X$ of atoms that unify
  wrt.\ $eq$ (called a \quo{cluster}),
  \eqref{eqCluster} defines
  a relation that contains $\mifactorcluster(s_1,s_2)$
  for all $s_1,s_2 \ins X$ where $s_2 \lts s_1$.
  \eqref{eqClusterNonlargest}
  represents all constants in all clusters
  that can be factored with a smaller element.
  \eqref{eqClusterFactorVia} uses constants that
  have no such smaller element as representatives
  and defines $\mifactorvia(s,s')$
  for all $s, s' \ins X$ where $s'$ is the smallest element
  of the respective cluster $X$.
  Finally, every atom $s$ that was unified
  with a representative $s'$ in $\mifactorvia(s,s')$
  is represented as factored $\mifactor(s)$
  by \eqref{eqClusterFactor}
  and those atoms that are neither factored
  nor inferred are defined as $\miabduce(s)$
  by \eqref{eqClusterAbduce}.
  Hence, in the answer set $I' \ins \AS(P_\bwda(A))$,
  all atoms $s \ins H$ that
  (i) are not back-chained on, i.e., are not marked as inferred,
  and that
  (ii) can be unified with a lexicographically smaller atom $s'$,
  are marked as $\mifactor(s) \ins I'$,
  and the factoring edge $(s,s') \ins E(G)$
  is represented as $\mifactorvia(s,s') \ins I'$.
  Those atoms $s \ins H$ that are neither factored nor inferred
  are marked as $\miabduce(s) \ins I'$.
  As $I \ins \AS(P_\mi{gpeq}(A))$ is in 1-1 correspondence
  with proof graphs based on backchaining
  and some equivalence relation,
  and $P_c$ adds to that the representation
  of factored and abduced atoms
  ($\mifactor(\cdot)$ and $\miabduce(\cdot)$)
  and factoring edges $\mifactorvia(\cdot,\cdot)$,
  $I' \ins \AS(P_\bwda(A))$ is in 1-1 correspondence
  with proof graphs based on back-chaining and unification,
  and the equivalence relation that supports this unification.
}%

\subsection{\fwda: Guess Abduced Atoms, Forward Inference, Check Goal}
\label{secFwda}

The previous encodings are based on explicitly representing proof graphs.
We next describe an encoding that is more in the spirit of the
generate-define-test paradigm of Answer Set Programming \cite{Lifschitz2002}:
we again propagate potentially interesting truth values,
however we do not keep track of the inferences.
We guess which of these potentially interesting truth values is abduced or factored,
and use another rewriting of the axioms to
reproduce their semantics,
i.e., we define that the head of an axiom is true
if all its bodies are true.
Finally we check that all goals become true.
For example axiom \eqref{eqEx4} is translated into
\begin{eqnarray}
  &&\miinfer(c(\miis,X,\midepressed)) \lars
    \mitrue(c(\miimportantfor,Y,X)), \mitrue(c(\miis,Y,\midead)).\label{eqEx4Forward} \\
  &&\mipot(c(\miimportantfor,Y,X)) \lars
    \mipot(c(\miis,X,\midepressed)), Y \eqs s(r_1,y,X). \label{eqEx4Pot1} \\
  &&\mipot(c(\miis,Y,\midead)) \lars
    \mipot(c(\miis,X,\midepressed)), Y \eqs s(r_1,y,X). \label{eqEx4Pot2}
\end{eqnarray}
where $r_1$ again is a unique identifier for axiom \eqref{eqEx4}.

We guess potential atoms as factored or abduced,
define truth from factoring, abducing, and inference,
and require that goals are true.
\begin{align}
  \{\ \mifai(X) : \mipot(X)\ \} \lars&. \\
  \mitrue(X) \lars& \mifai(X). \\
  \mitrue(X) \lars& \miinfer(X). \\
  \lars& \migoal(A), \naf\ \mitrue(A).
\end{align}
This realizes abduction in the classical generate-define-test way.
The only thing missing is factoring to determine
which atoms actually need to be abduced.
For that we add the following rule
to define which atoms are factored or abduced.
\begin{align}
  \mifa(X) \lars \mifai(X), \naf\ \miinfer(X).
\end{align}

We complete the encoding by cluster factoring rules
\eqref{eqCluster}--\eqref{eqClusterAbduce}
and common rules \eqref{eqPotential}--\eqref{eqEqTransitive}.

Because we do not have an explicit representation of the proof tree,
only the \cardobj\ metric is applicable.
Moreover, we cannot factor with inferred atoms
as there is no way
to rule out circular inference,
hence we only study the most restricted factoring variant.

For readability we gave the rewriting as an example.
Formally, an axiom of form \eqref{eqAx} is rewritten into
\begin{align*}
  \miinfer(c(q,Y_1,\ldots,Y_m)) &\lars
    \mitrue(c(p_1,X^1_1,\ldots,X^1_{k_1})),
    \ldots,
    \mitrue(c(p_r,X^r_1,\ldots,X^r_{k_r})). \\
  \mipot(c(p_i,X^i_1,\ldots,X^i_{k_1})) &\lars
      Z_1 \eqs s(a,1,Y_1,\ldots,Y_m), \ldots,
       Z_v \eqs s(a,v,Y_1,\ldots,Y_m), \\
  &\quad \mipot(c(q,Y_1,\ldots,Y_m)).
         \qquad\qquad\qquad\qquad\qquad\qquad \text{ for } i \ins \{ 1, \ldots, r \}
\end{align*}
where $a$ is a unique identifier for that particular axiom
and $Z_1,\ldots,Z_v \eqs \cX \setminuss \cY$.
Note that this means that the resulting rules will all be safe.

We do not prove correctness of this encoding
as it is only applicable to \cardobj\ and
does not have good performance
compared with other encodings.

\subsection{Encodings for Preferred Solutions}
\label{secPreferences}

We next describe program modules
that realize objective functions
when we add them to the previously given encodings.

\paragraph{Cardinality Minimality.}
For realizing objective \cardobj\ we use the following weak constraint.
\begin{align}
  \label{eqCardObj}
  \wlar \miabduce(P).\quad [1@1,P]
\end{align}

\paragraph{Coherence Metric.}
For \accelobj\ we represent
which nodes are reachable from which goal node.
\begin{align}
  \label{eqCohReachGoal}
  \mireach(P,P) &\lars \migoal(P). \\
  \label{eqCohReachInfer}
  \mireach(Q,\mi{From}) &\lars \mireach(P,\mi{From}), \miinfervia(R,P), \miinferenceneeds(P,R,Q). \\
  \label{eqCohReachFactor}
  \mireach(Q,\mi{From}) &\lars \mireach(P,\mi{From}), \mifactorvia(P,Q).
\end{align}
We represent pairs of distinct goal atoms that
have a common reachable atom,
and we create a weak constraint
that incurs a cost corresponding to pairs of goal atoms
without a common reachable atom.
\begin{align}
  \label{eqCohReachFromBoth}
  \mireachfromboth(P,Q) &\lars \migoal(P), \migoal(Q), P \lts Q, \mireach(N,P), \mireach(N,Q). \\
  &\wlar \migoal(P), \migoal(Q), P \lts Q, \naf\ \mireachfromboth(P,Q).\quad [1@1,P,Q]
  \label{eqCohObj}
\end{align}

\paragraph{Weighted Abduction.}
For realizing \waobj\
we represent potential costs as an integers.
We seed cost with $\$ 100$ for goal atom assumption cost.
\begin{align}
  \label{eqWAGoal}
  \mipcost(P,100) &\lars \migoal(P).
\end{align}
As common practice for applying \waobj\ to \accel,
we realize axiom costs such that body cost factors
sum up to $1.2$.
For that we require for each axiom $R$ a fact
$\minumberofbodies(R,N)$ to be defined
such that $N$ is the number of body atoms of $R$.
We also require a minimum cost of $1$
which prevents spurious very deep proof trees from being optimal
due to cost $0$ at abduced atoms.
\begin{align}
  \label{eqWAInfer}
  \mipcost(Q,Mc) \lars&
    \miinfervia(R,P), \miinferenceneeds(P,R,Q), \\
    &Mc \eqs \mi{\#max}\ \{\ (C*6/5)/N\ ;\ 1\ \},
      \mipcost(P,C), \minumberofbodies(R,N). \notag
\end{align}
These computations are handled during instantiation.
They can be generalized to assumption weights
that are individually given for each axiom as facts,
without causing a change the rest of the encoding.

We propagate cost across factoring edges,
define cost to be the minimum cost found at all abduced atoms,
and minimize that cost using a weak constraint.
\begin{align}
  \label{eqWAFactor}
  \mipcost(Q,C) &\lars \mifactorvia(P,Q), \mipcost(P,C). \\
  \label{eqWACost}
  \micost(P,C) &\lars \miabduce(P),\
    C \eqs \mi{\#min}\ \{\ Ic : \mipcost(P,Ic)\ \}. \\
  &\wlar \micost(P,C).\quad [C@1,P]
  \label{eqWAObj}
\end{align}

\mytableEncodings{tbp}
\begin{proposition}
  \label{thmObjectiveEncodings}
  Let $P_\cardobj = \{$\eqref{eqCardObj}$\}$,
  $P_\accelobj = \{$\eqref{eqCohReachGoal}--\eqref{eqCohObj}$\}$,
  and $P_\waobj = \{$\eqref{eqWAGoal}--\eqref{eqWAObj}$\}$.
  Then the cost of answer sets $I \in \AS(P_\bwda(A) \cup P_\cardobj)$,\
  $I \in \AS(P_\bwda(A) \cup P_\accelobj)$, and
  $I \in \AS(P_\bwda(A) \,{\cup}$ $P_\waobj)$
  is the objective function $\cardobj(G)$, $\accelobj(G)$, and $\waobj(G)$,
  respectively, of the proof graph $G$ represented in $I$,
  where for \waobj\ costs are rounded to integers and at least $1$.
\end{proposition}%
\newcommand{\myProofThmObjectiveEncodings}{
  (\cardobj)
  \eqref{eqCardObj} causes cost $|\{ a \mids \miabduce(a) \ins I \}|$
  and $\miabduce(a) \ins I$ iff $a$ is neither inferred nor factored.
  As being abduced is equivalent with the absence of edges $\mifactorvia(a,a') \ins I$
  and edges $\miinfervia(r,a) \ins I$, \eqref{eqCardObj} produces cost $\cardobj(G)$
  for an answer set $I$ representing $G$.

  (\accelobj)
  \eqref{eqCohReachGoal} seeds a definition of reachability
  from goal nodes in predicate $\mireach$,
  \eqref{eqCohReachInfer} defines reachability
  across those inference
  edges that correspond with inferences actually done
  in the proof graph represented in $I$,
  and \eqref{eqCohReachFactor} defines reachability
  across factoring edges.
  As a result $\mireach(a,o)$ is true for atom
  $a \ins H$ and observation $o \ins O$
  iff $o$ is reachable from $a$ in $G$.
  \eqref{eqCohReachFromBoth} defines
  $\mireachfromboth(o,o') \ins I$
  iff $o, o' \ins O$, $o \lts o'$,
  and there is some atom $a \ins H$ such that
  a is reachable from $o$ and $o'$.
  The if direction is ensured by rule satisfaction,
  the only if direction is ensured by answer set minimality.
  Finally, weak constraint \eqref{eqCohObj}
  attaches cost $1$ for each distinct pair $o,o'$
  of observation nodes where
  no node reachable from both $o$ and $o'$ exists in $G$.
  This exactly corresponds to the definition of \accelobj.

  (\waobj)
  \eqref{eqWAGoal} defines potential cost of $100$
  for objective nodes $o \ins O$.
  \eqref{eqWAInfer} defines potential cost of
  $1.2 \cdot c/n$ for each body atom of back-chaining,
  given that the back-chained atom had potential cost $c$
  and was back-chained over a rule with $n$ body atoms.
  \eqref{eqWAFactor} defines that for atoms $p,q$
  where $p$ was factored with $q$,
  $q$ obtains all potential costs from $p$.
  Potential cost includes all costs
  obtained from reachable nodes,
  including the minimum cost in case of factoring.
  Therefore, the minimum costs propagated for unification
  as described in Def.~\ref{defGoalFun}
  is represented as potential cost, along with bigger costs.
  Rule \eqref{eqWACost} represents for each abduced atom,
  i.e., for each $a \ins A(G)$ for $G$ represented in $I$,
  the minimum over all potential costs of $a$.
  Hence, $\micost(p,c) \ins I$ iff $p \ins A(G)$,
  and $c$ is the cost of abduced atom $p \ins H$
  according to \waobj.
  \eqref{eqWAObj}
  sums up costs of distinct atoms $p$,
  hence cost $\waobj(G)$ is assigned to $I$.
}%

\rev{}{\subsection{Summary}
Table~\ref{tblEncodings} gives an overview of our encodings.
Encodings in the \bwd\ family deterministically represent
the maximal potential proof graph
and all its inferences, while \fwda\ represents
only atoms in the hypothesis.
Justification of atoms in the proof graph is ensured
from goals to abducibles (backward)
in the \bwd\ encodings, and from abduced atoms to goals
(forward) in \fwda.
The type of justification of goals and atoms in the hypothesis
is nondeterministically guessed as one of three classes by \bwdg
and one of two classes by \bwdai\ and \bwda.
\fwda\ guesses truth of abduced or factored atoms (two classes).
Factoring is canonicalized to various extent,
and acyclicity of factoring is implicitly ensured
in \bwda\ and \fwda\ but is encoded explicitly
in other encodings.}

\section{Extensions}

The ASP encodings given so far are realized in pure ASP-Core-2 \rev{}{syntax}
and do not require additional features specific to particular
solver tools.
However, these encodings have two drawbacks:
they can represent only acyclic theories,
and they have performance issues related to
the size of instantiation
of the transitivity constraint for $\mieq$.
We next formalize two extensions of these encodings
and \rev{}{describe} their computational realization.

In Section~\ref{secFlexibleValueInvention}
we introduce Flexible Value Invention
for fine-grained control of Skolemization,
which makes our encodings applicable to cyclic theories.
In Section~\ref{secOnDemand} we show
how to replace certain constraints in our encodings
with lazy variants to reduce grounding size
and potentially improve evaluation performance.
Both extensions are described formally
in the \hex\ formalism.
Section~\ref{secImpl} discusses how we realized
these extensions using the Python library of
\rev{\clasp\ and \gringo}{\clingo}.

\subsection{Flexible Value Invention for Cyclic Knowledge Bases}
\label{secFlexibleValueInvention}

The encodings in Section~\ref{secMainEncodings}
assume that the knowledge base is acyclic,
which ensures finite proof trees
and a finite instantiation of our ASP encodings
in the presence of Skolemization
with Uninterpreted Function terms as done
in our axiom rewritings.

\begin{example}
  As an example of a cyclic knowledge base consider the following two axioms
  \begin{align}
    p(A,b) &\,\lax\, q(A,C), t(C,b). \tag{$r_1$} \\
    t(D,b) &\,\lax\, p(D,b). \tag{$r_2$}
  \end{align}
  where a goal of $p(a,b)$ yields the following infinite
  backward chaining instantiation of axioms in the proof tree
  \begin{align}
    p(a,b) &\,\lax\,
      q(a,s(r_1,``C",a)), t(s(r_1,``C",a),b). \tag{via $r_1$}\\
    t(s(r_1,``C",a),b) &\,\lax\,
      p(s(r_1,``C",a),b). \tag{via $r_2$} \\
    p(s(r_1,``C",a),b) &\,\lax\,
      q(s(r_1,``C",a),s(r_1,``C",s(r_1,``C",a))), \notag \\
      &\hphantom{\,\lax\,} t(s(r_1,``C",s(r_1,``C",a)),b). \tag{via $r_1$} \\
    t(s(r_1,``C",s(r_1,``C",a)),b) &\,\lax\,
      p(s(r_1,``C",s(r_1,``C",a)),b). \tag{via $r_2$} \\
    &\ \:\:\vdots \notag
  \end{align}
  where $C$ is first skolemized as $s(r_1,``C",a)$
  (see~\eqref{eqEx4rewriteHead})
  but then used again to back-chain over the first axiom
  which leads to another Skolemization.
  This leads to undecidability
  as we cannot know when we have generated
  `enough' distinct variables to find the optimal solution.
  \hfill$\square$
\end{example}

The \accel\ benchmark is described as being acyclic \cite{Ng1992kr}
however it contains one cyclic axiom
and this contains a comment that suggests that the respective
axiom has been added after publication of \cite{Ng1992kr}.
To evaluate \accel, or any cyclic theory with our encodings,
we therefore need to exclude axioms to break cycles,
or infinite instantiations will occur.
However, in knowledge representation,
knowledge is sometimes naturally expressed
in cyclic axioms, and we would like to handle
such knowledge bases.
In particular the cyclic axioms in \accel\ are required
to obtain correct solutions for some instances,
so we do not want to dismiss such axioms.

We next use external atoms
instead of uninterpreted function symbols
to achieve a \emph{Flexible Value Invention}
where we can control when to block value invention.
\rev{}{By blocking certain value inventions,
we ensure a finite instantiation of our encoding
which thereby allows computation of optimal solutions.
If we do \emph{not} use cyclic axioms and limit value invention,
we obtain a subset of all acyclic proof graphs
and a sound approximation of the optimal solution.
If we use cyclic axioms, they extend the proof graph
in ways that are impossible with acyclic axioms.
The variations of Flexible Value Invention (shown in the following)
permit usage of cyclic axioms and
allow for controlling the trade-off between
instantiation size and distance from the optimal solution.}

For \rev{that}{Flexible Value Invention}, we \rev{}{first }outsource value invention into an external atom
$\amp{skolem}$.
Formally, instead of skolemizing variables $Z_{\rev{i}{v}} \ins \cX \setminus \cY$ \rev{}{in the rewriting \eqref{eqFormalBwd} }as
\begin{align}
Z_{\rev{i}{v}} \eqs s(a,\rev{i}{v},Y_1,\ldots,Y_m) \label{eqFunSkolem}
\intertext{%
where $a$ is the axiom identifier\rev{}{, $v$ is the variable index, and $Y_1,\ldots,Y_m$ are head variables}, we use
}
\ext{skolem}{a,\rev{i}{v},Y_1,\ldots,Y_m}{Z_{\rev{i}{v}}}. \label{eqExtSkolem}
\end{align}

\begin{example}
  Instead of \eqref{eqEx4rewriteHead}
  in the \bwd\ encodings \rev{}{(Example~\ref{exUFSkolemizationRewriting})},
  we rewrite into the rule
  \begin{align}
    &\mimayinfervia(r_1,c(\miis,X,\midepressed),l(Y))
      \lars \notag \\
    &\hspace*{16em} \mipot(c(\miis,X,\midepressed)),
      \ext{skolem}{r_1,``Y",X}{Y} \label{eqEx4rewriteHeadSkolem}
  \intertext{
  and instead of \eqref{eqEx4Pot1} and \eqref{eqEx4Pot2} in the \fwda\ encoding
  we rewrite the body elements of the axiom into
  }
    &\mipot(c(\miimportantfor,Y,X)) \lars
      \mipot(c(\miis,X,\midepressed)),
      \ext{skolem}{r_1,``Y",X}{Y}. \notag \\
    &\mipot(c(\miis,Y,\midead)) \lars
      \mipot(c(\miis,X,\midepressed)),
      \ext{skolem}{r_1,``Y",X}{Y}. \notag
  \end{align}
  \hfill$\square$
\end{example}

\rev{}{This way we outsource Skolemization,
i.e., building a new unique constant term $Z_v$ from terms
$a,v,Y_1,\ldots,Y_m$ --- or the decision not to build such a term ---
to an external computation.}
We next realize several Skolemization methods that limit value invention in
\rev{the external computation, and we outsource the necessary bookkeeping
to the external source.}{different ways.}

The original Skolemization with uninterpreted functions
can be emulated by defining \extfun{skolem} as
\begin{align*}
  \extsem{sk^\infty}{I,R,V,Y_1,\ldots,Y_m,Z} = 1
  \text{ iff } Z = s(R,V,Y_1,\ldots,Y_m).
\end{align*}
This shows that we can still express the original
Skolemization (without guaranteeing decidability).

\begin{example}[continued]
\rev{}{The external atom $\ext{sk^\infty}{r_1,``Y",X}{Y}$
is true iff $Y \eqs s(r_1,``Y",X)$.
This means that instantiating
\eqref{eqEx4rewriteHead}, which contains an uninterpreted function,
and instantiating \eqref{eqEx4rewriteHeadSkolem},
which contains an external computation,
will create the same ground Skolem terms.}
\hfill$\square$
\end{example}

A simple way for ensuring termination is
the following function.
\begin{align}
  \label{eqSkolemP1}
  \extsem{sk^{P^1}}{I,R,V,Y_1,\ldots,Y_m,Z} = 1
  \text{ iff }
  \left\{ \begin{array}{@{~}l@{}}
    Z = s(R,V,Y_1,\ldots,Y_m) \\
    \text{ and no $Y_i$, $1 \les i \les m$, is of the form $s(\cdot,\cdot,\cdots)$.}
  \end{array}\right.
\end{align}
This prevents value invention if any of the terms
$Y_1,\ldots,Y_m$ is an invented value,
which is a very restrictive criterion:
it blocks all value invention where at least one parent
is an invented value.

\begin{example}[continued]
\rev{}{The external atom $\ext{sk^{P^1}}{r_1,``Y",X}{Y}$
is true if $Y \eqs s(r_1,``Y",X)$ and $X$ is not a term of form $s(\cdots\!)$.
Instantiating rule \eqref{eqEx4rewriteHeadSkolem}
with $X \eqs m$ then yields a single external atom
$\ext{sk^{P^1}}{r_1,``Y",m}{s(r_1,``Y",m)}$ which evaluates to true.
Therefore, the rule head is instantiated as
$\mimayinfervia(r_1,c(\miis,m,\midepressed)$, $l(s(r_1,``Y",m)))$.
Assume that we have an additional axiom
which contains $\miis(X,\midepressed)$ in the head
and $\miis(X,\midead)$ or $\miimportantfor(X,Z)$ in the body.
Such an axiom allows cyclic back-chaining over $\miis(X,\midepressed)$,
which yields another instantiation of \eqref{eqEx4rewriteHeadSkolem}
with body literals
$\mipot(c(\miis,s(r_1,``Y"$, $m),\midepressed))$ and
$\ext{sk^{P^1}}{r_1,``Y",s(r_1,``Y",m)}{Y}$.
In this case the external atom will not be true
for any ground term $Y$:
it blocks Skolemization as the input term
is already a Skolem term.
Without this blocking (e.g., with $\amp{sk^\infty}$)
we would obtain an infinite instantiation.}
\hfill$\square$
\end{example}

We can extend $\extfun{sk^{P^1}}$ to block value invention only
if some grandparent is an invented value.
\begin{align}
  \label{eqSkolemP2}
  \extsem{sk^{P^2}}{I,R,V,Y_1,\ldots,Y_m,Z} = 1
  \text{ iff }
  \left\{ \begin{array}{@{~}l@{}}
    Z = s(R,V,Y_1,\ldots,Y_m) \\
    \text{ and no $Y_i$, $1 \les i \les m$, is of the form $s(\cdot,\cdot,U_1,\ldots,U_k)$} \\
    \text{ \quad with some $U_j$, $1 \les j \les k$,
      of the form $s(\cdot,\cdot,\cdots)$.}
  \end{array}\right.
\end{align}
This can be further generalized to $\extfun{sk^{P^i}}$,
where $i \ins \{ 1, 2, \ldots \}$,
indicates that in the $i{-}1$-th nesting level of terms
$Y_i$, $1 \les i \les m$, terms must not be invented values.
$P^1$ corresponds to the method used in \henry\ for achieving
termination (Naoya Inoue 2015, personal communication).
\smallskip

External oracle functions $\extfun{sk^{P^i}}$, $1 \les i$,
guarantee finite instantiation of cyclic abduction problems.
\begin{proposition}
  \label{thmSkolemPFiniteness}
  Given a cyclic or acyclic abduction instance $A \eqs (B,O,S)$,
  let ${P_\bwda}^{P^i}(A)$
  be the program $P_\bwda(A)$
  after replacing all body atoms of form \eqref{eqFunSkolem}
  by body atoms of form \eqref{eqExtSkolem}
  where $\amp{skolem} \eqs \amp{sk^{P^i}}$,
  then $\grnd({P_\bwda}^{P^i}(A))$ is finite
  for $i \ins \{ 1, 2, \ldots \}$.
\end{proposition}%
\newcommand{\myProofThmSkolemPFiniteness}{
  ${P_\bwda}^{P^i}(A)$ is finite and contains only safe rules,
  therefore the only source of infinite instantiation
  can be the instantiation of terms of unlimited depth.
  Except for the first rule in the axiom rewriting \eqref{eqFormalBwd},
  all rules have heads with term nesting level
  equal or lower than in the body,
  hence the only rule that can generate terms of unlimited depth is
  the first in \eqref{eqFormalBwd}.
  In that rule, term $l(\cdots)$ is created,
  but it is only used in the other rewritten rules in \eqref{eqFormalBwd}
  where only the arguments of $l(\cdots)$ are used,
  and in \eqref{eqInferViaWhat}, where this term is discarded,
  therefore $l(\cdots)$ cannot be infinitely nested.
  The only source of terms of infinite depth is the external atom
  of form $\ext{sk^{P^i}}{a,i,Y_1,\ldots,Y_m}{Z_i}$
  and it causes infinite instantiation
  only if $\extfun{sk^{P^i}}(I,R,V,Y_1,\ldots,Y_m,Z)$
  is true for infinitely many distinct terms $Z$.
  $\extfun{sk^{P^i}}$ is $1$
  only if no input $Y_i$, $1 \les i \les m$,
  has a subterm at nesting level $i-1$
  that is of form $s(\cdots)$,
  and $R,V$ can be a finite number of constants from $P_\bwda^{P^1}(A)$.
  Moreover there is a finite number of terms that can be built from
  constants in ${P_\bwda}^{P^i}(A)$ with function symbol $s$,
  and not having a subterm of form $s(\cdots)$ below nesting level $i-1$.
  Hence there is a finite number of tuples
  $R,V,Y_1,\ldots,Y_m$
  for which $\extfun{sk^{P^i}}$ evaluates to true.
  As $Z$ depends on $R,V,Y_1,\ldots,Y_m$,
  $\extfun{sk^{P^i}}$ evaluates to true
  for a finite number of tuples
  $R,V,Y_1,\ldots,Y_m,Z$
  and instantiation is finite
  with respect to a given program ${P_\bwda}^{P^i}(A)$.
}%
\begin{proof}
  \myProofThmSkolemPFiniteness
\end{proof}

Limiting value invention this way is not the only way:
nontermination of value invention
always involves a certain rule
and a certain existential variable of that rule
being instantiated over and over again in a cycle.
Therefore, we next formulate
an external Skolemization oracle
that blocks Skolemization only if a child term
was generated for the same rule and variable.
\begin{align}
  \label{eqSkolemGen}
  \extsem{sk^{G^1}}{I,R,V,Y_1,\ldots,Y_m,Z} = 1
  \text{ iff }
  \left\{ \begin{array}{@{~}l@{}}
    Z = s(R,V,Y_1,\ldots,Y_m) \text{ and} \\
    \text{no $Y_i$, $1 \les i \les m$, has a sub-term of form $s(R,V,\cdots)$.}
  \end{array}\right.
\end{align}
This function also ensures finite instantiation.
\begin{proposition}
  \label{thmSkolemGFiniteness}
  Given a cyclic or acyclic abduction instance $A \eqs (B,O,S)$,
  let ${P_\bwda}^{G^1}(A)$
  be the program $P_\bwda(A)$
  after replacing all body atoms of form \eqref{eqFunSkolem}
  by body atoms of form \eqref{eqExtSkolem}
  where $\amp{skolem} \eqs \amp{sk^{G^1}}$,
  then $\grnd({P_\bwda}^{G^1}(A))$ is finite
  for $i \ins \{ 1, 2, \ldots \}$.
\end{proposition}%
\newcommand{\myProofThmSkolemGFiniteness}{
  As in the proof of Proposition \ref{thmSkolemPFiniteness},
  the only reason for infinite instantiation can be
  the external atom.
  Assume towards a contradiction,
  that the instantiation is infinite.
  Then $\extfun{sk^{G^1}}(I,R,V,Y_1,\ldots,Y_m,Z)$ must be true
  for an infinite number of terms $Z$.
  As we have a finite number of constants in ${P_\bwda}^{G^1}(A)$
  and $Z$ is instantiated using this set of constants
  and function symbols of form $s(\cdots)$ with finite arity,
  for an infinite number of terms we require infinite nesting depth
  of terms of form $s(\cdots)$.
  As the set of possible tuples $(R,V)$ used as inputs of
  $\extfun{sk^{G^1}}$ is finite,
  we must repeat $(R,V)$ in some subterms of $Z$
  to reach an infinite amount of them.
  However $\extfun{sk^{G^1}}$ is false for such terms,
  contradiction.
}%
\begin{proof}
  \myProofThmSkolemGFiniteness
\end{proof}
As before, we can further generalize
to $\extfun{sk^{G^i}}$ where $i \ins  \{ 1, 2, \ldots\}$
indicates that a Skolem term may contain at most $i$ layers of sub-terms from the same rule and variable.

\subsection{On-Demand Constraints}
\label{secOnDemand}

\rev{}{In the set of rules common to all encodings (Section~\ref{secCommonRules}),}
we represented transitivity
of the equivalence relation $\mieq$
as a constraint \rev{}{\eqref{eqEqTransitive} }instead of using
\rev{a rule in \eqref{eqEqTransitive}
(i.e., shifting $\naf\ \mieq(A,C)$ into the head).}%
{the more commonly used rule}
\begin{align*}
  \mieq(A,C) \lars \mieq(A,B), \mieq(B,C), A \rev{\lts}{\neqs} B, B \rev{\lts}{\neqs} C, \rev{}{A \neqs C}.
\end{align*}
\rev{This way we can}{The formulation as a constraint allows us to} eliminate \eqref{eqEqTransitive} from the encoding
and lazily create only those instances of \eqref{eqEqTransitive}
that are violated during the search for the optimal solution.
Such \rev{}{a }lazy instantiation is easy for constraints
but \rev{difficult}{not supported} for rules
\rev{}{ in current solvers},
as adding a new rule changes the solver representation
(usually the Clark completion)
of all rules with the same head in the program.

Formally we represent lazy constraints in \hex\
(cf.~\cite[2.3.1]{Eiter2015hex})
by replacing \eqref{eqEqTransitive} with a constraint
\begin{align}
  &\lars \naf\ \ext{transitive}{\mieq}{}. \label{eqEqOndemand} \\
\intertext{
where the external computation oracle is defined as follows:
}
  &\extsem{transitive}{I,p} = 1 \text{ iff }
  \text{the relation $\{ (A,B) \mids p(A,B) \ins I\}$ is transitive.} \label{eqExtTransitive}
\end{align}
Moreover, if we find a transitivity violation,
i.e., a new triple $(A,B,C)$ such that
$\{ p(A,B), p(B,C) \} \subseteqs I$
and $p(A,C) \notins I$,
we add a new nogood into the solver
that prevents $p(A,B) \lands p(B,C) \lands \neg p(A,C)$
for future answer set candidates.

Similarly, we can ensure that $\mibelow$ is a partial order,
by replacing \eqref{eqBelow2} with a guess of the
relevant part of the extension of predicate $\mibelow$%
\footnote{I.e., the part used in \eqref{eqGuessFactorInfer} and \eqref{eqIAClusterFactorInfer}.}
as follows
\begin{align*}
  \{\ \mibelow(Q,P)\ \} \lars \mifai(P), \miinfer(Q).
\end{align*}
and require acyclicity of $\mibelow$
using a constraint of form \eqref{eqEqOndemand}
with external atom $\ext{acyclic}{\mibelow}{}$,
an oracle function that is true
iff relation $\{ (A,B) \mids p(A,B) \ins I \}$
is acyclic, and nogood generation
for all basic cycles in that relation.

\subsection{Implementation}
\label{secImpl}

We formulated
Flexible Value Invention and On-Demand Constraints
using \hex\ as a formal framework.
In preliminary experiments we identified a performance problem
in the \dlvhex\ solver that was not possible to fix easily,
therefore we decided to realize the extensions using
the Python libraries of \gringo\ and \clasp\
\cite{Gebser2014clingorepext}.
This posed additional challenges that we discuss next.

\paragraph{Flexible Skolemization.}
Flexible Skolemization can be realized purely
\rev{in grounding}{during instantiation}
by replacing external atoms of the form
$\ext{skolem}{a,i,Y_1,\ldots,Y_m}{Z_i}$
by the expression
$Z_i = @\mi{skolem}(a,i$, $Y_1,\ldots,Y_m)$
and implementing a Python function {\tt skolem}
that generates constants
according to the various semantic definitions
for limited value invention that are described
in Section~\ref{secFlexibleValueInvention}.

Note that we can only handle this kind of external atoms
in grounding because the value of the oracle function
$\extfun{skolem}$ does not depend on the interpretation $I$.

\paragraph{On-demand Constraints.}
Different from Flexible Skolemization,
on-demand constraints are handled during solving.
For that, \clasp\ provides an interface for
registering a callback function which receives answer set candidates.
In that callback we can add nogoods to the solver.
However, answer set enumeration modes
of the current Python API of \clasp\ do not
work well together with this idea:
either we enumerate models of increasing quality,
or we enumerate models without using the objective function.
In the former case an on-demand constraint
can invalidate the first model of optimal quality,
which causes no further answer set candidates to be found,
(there are no better ones).
The latter case is a blind search for better solutions
which is prohibitively slow.

\rev{We realized an algorithm that}%
{To realize on-demand constraints with reasonable efficiency,
we created algorithm \textsc{Find\-Optimal\-Model} which}
first finds an optimistic bound for the objective function
and then backtracks to worse bounds
using on-demand constraints.
This algorithm is of interest only until the \clasp\ API
supports changing enumeration mode or objective bound
from within the callback,
hence we show details \rev{in Appendix~\ref{secOnDemandAppendix}}%
{only in the appendix}.

\paragraph{Global Constraints of \accel.}
A final implementation aspect is the realization
of global constraints of the \accel\ benchmark.

Assumption nogoods and unique-slot constraints can be
represented uniformly for all encodings
in ASP constraints.
We here show the encoding strategy by means of examples.
Assumption nogoods as exemplified in \eqref{eqExAssumption}
are encoded in ASP as
\begin{align*}
  \lars&
    \miabduce(c(\mi{go\_step},S,G_1)),
    \miabduce(c(\mi{goer},G_2,P)),
    \mieq(G_1,G_2)
\end{align*}
where we take into account term equivalence.

Unique slot axioms as exemplified in \eqref{eqExUniqueSlot}
are encoded in ASP as
\begin{align*}
  \lars&
    \mitrue(c(\mi{goer},G_1,P_1)),
    \mitrue(c(\mi{goer},G_2,P_2)),
    \mieq(G_1,G_2),
    P_1 \lts P_2,
    \naf\ \mieq(P_1,P_2)
\end{align*}
where we take into account term equivalence both for
the entity that must be the same to violate the constraint
($G_1$, $G_2$), and for the entity
that is enforced to be unique,
i.e., must not be the same to violate the constraint ($P_1$, $P_2$).
Note that condition $P_1 \lts P_2$ achieves symmetry breaking
during instantiation.

\section{Experimental Evaluation}
\label{secExperiments}
We evaluated the above encodings, on-demand constraints, and flexible value invention using the \accel\ benchmark described in Section~\ref{secAccel}.
The encodings and instances we used in experiments are available online.%
\footnote{\url{https://bitbucket.org/knowlp/asp-fo-abduction}}
The benchmarks were performed on a computer with 48 GB RAM
and two Intel E5-2630 CPUs (total 16 cores)
using Ubuntu 14.04.
As solvers we used the Python API of \clingo%
\footnoteremember{fnClingo}{\rev{}{\url{http://potassco.sourceforge.net/}}}
4.5.\rev{0}{4}%
\footnoteremember{fnPatch}{\rev{}{Including a patch that will be contained in future versions and eliminates a bug with aggregates.}}
\rev{}{\cite{Gebser2014clingorepext}}
to implement \rev{O}{o}n-demand constraints and \rev{F}{f}lexible Skolemization as described in Section~\ref{secImpl},
and we tested pure ASP encodings also with \gringo%
\rev{}{\footnoterecall{fnClingo}} 4.5.\rev{0}{4}%
\footnoterecall{fnPatch}
\cite{Gebser2011gringo3} as grounder
and both solvers \clasp%
\rev{}{\footnoterecall{fnClingo}} \rev{4.5.0}{3.1.4} \cite{Gebser2015clasp3}
and \wasp%
\footnote{\url{https://github.com/alviano/wasp/}\rev{version~f9d436}{}}%
\rev{}{\ version~f9d436}
\cite{Alviano2015wasp}. %
We also make experiments to compare with
state-of-the-art approaches
\henry%
\footnote{\url{https://github.com/naoya-i/henry-n700}\rev{~version 4b0d900}{}}%
\rev{}{~version 4b0d900 }%
\cite{Inoue2013}
and its successor \phillip%
\footnote{\url{https://github.com/kazeto/phillip}\rev{~version 5612b13}{}}%
\rev{}{~version 5612b13 }%
\cite{Yamamoto2015}.
Each run was limited to 5 minutes and 5 GB RAM,
\htcondor\ was used as a job scheduling system,
each run was repeated 5 times
and no more than 8 jobs were running simultaneously.
For \clasp\ we used the setting \verb|--configuration=crafty|
which turned out to be superior to all other preset configurations of \clasp.
For \wasp\ we \rev{did not use any specific configuration.}%
{used the default configuration (core-based OLL algorithm)
which performs equal or better compared with other settings
(note in particular, that configurations}
\verb|basic|, \verb|mgd|, and \verb|opt|
\rev{}{for option}
\verb|--weakconstraints-algorithm|
\rev{}{perform clearly worse than the default).}

\mytableGeneral{tb}
In the following tables,
columns Opt, To, and Mo \rev{(if shown) }{}give the number of instances
for which an optimal solution was found,
the timeout was reached,
and the memory limit was exceeded, respectively.
These numbers are summed over instances and averaged over runs.
Columns \myt\ and \mym\ show time (seconds) and memory (MB) requirement,
averaged over instances and runs.

The remaining columns show detailed diagnostics of the solver and objective function, where provided by the respective tool.
\mygrd\ and \myslv\ give grounding and solving time,
respectively,
as reported by running first \gringo\ and then
\clasp\ or \wasp.
In experiments with the Python implementation,
\mygrd\ includes solver preprocessing
which cannot be separated from grounding in \clasp\ API,
and \myslv\ contains pure search time.
Further metrics are only available if an optimal solution
could be found and proved as optimal:
\myobj\ shows the objective function,
\myodc\ the number of on-demand constraints,
\myskc\ the number of created Skolem constants,
\mychc/\myconf\ the number of choices and conflicts encountered,
and
\myrul\ the number of rules in the program.
To permit a meaningful comparison of these values,
we average them over the \rev{20}{17} easiest instances.
We only show averages if all runs found the optimum solution,
otherwise we write \quo{*}.
For large numbers, K and M abbreviate $10^3$ and $10^6$.

\rev{We will not numerically compare our results
with preliminary results obtained in \cite{Schuller2015rcra},
because we extended the functionality
and improved performance in all cases.}%
{Preliminary encodings in \cite{Schuller2015rcra}
were able to represent acyclic theories and therefore only suitable for objective \cardobj.
Equivalence was represented by a relation between terms
and representative terms in the respective equivalence class
(\textsc{BackCh}) or not at all (\textsc{Simpl}).
We will not make numerical comparisons
as the performance of \textsc{BackCh} and \textsc{Simpl}
is significantly worse than the performance of encodings
in this work in all cases
(in particular memory is exhausted in more than 50\% of instances with \clasp).}

\paragraph{Basic ASP Encodings.}
Table~\ref{tblGeneral} shows experimental results for\rev{ comparing}{}
encodings \bwdg, \bwdai, \bwda, and \fwda\ for
objectives \cardobj, \accelobj, and \waobj\ using
\gringo\ for grounding and \clasp\ (C) or \wasp\ (W)
for solving.
For \waobj\ we also compare with \phillip\ (P).
\bwdg\ \rev{clearly }{}performs worst
with respect to all metrics%
\rev{}{, it performs significantly worse than \bwdai\ with
\clasp, while it performs just a bit below \bwdai\ with \wasp}.
\bwda\ performs best with respect to all metrics,
\rev{except}{and} for \cardobj\rev{\ where}{,} \wasp\ performs
nearly the same \rev{as}{with} the \fwda\ encoding.

{\ifrevisionmarkers\color{blue}\fi
Comparing \clasp\ and \wasp\ shows that
\wasp\ is able to find the optimal solution
faster than \clasp\ except in cases
where \wasp\ exceeds the 5GB memory
limit, which happens mainly for the encodings
containing a higher amount of guesses (\bwdg, \bwdai).
Another difference is the number of choices
and conflicts: \wasp\ generates fewer choices
for \cardobj\ and \accelobj,
but more choices for \waobj, moreover
\wasp\ often generates more conflicts.
These differences on the same encoding
can be explained by different ASP optimization algorithms:
for \clasp\ the used (default) configuration BB
is based on adding constraints
to a relaxation of the instance,
while for \wasp\ the used (default) configuration OLL
is based on unsatisfiable cores
(for a discussion of optimization approaches, see
\cite{Alviano2015waspoptimum}).
}

Due to its unfavorable performance,
we omit results for \bwdg\ in the following.

The \henry\ solver realizes \waobj\ and on the \accel\ benchmark
results of around 10~sec per instance have been reported \cite{Inoue2013}.
\phillip\ is the successor of \henry\ and adds heuristics
for a partial instantiation of the most relevant portions of the proof graph
\cite{Yamamoto2015}.
We experimented with \henry\ and \phillip\ on
\rev{a rewriting of \accel\ that is close
to the original \accel\ knowledge base.}%
{the original \accel\ knowledge base.}
Unfortunately we were not able to reproduce the results reported for \henry:
as shown in the table,
all 50 instances timed out,
moreover an ILP solution was found only
for 4 instances and the costs of these solutions
where more than 25\% above the optimal result.
As we had problems with the Gurobi license,
we used the open-source \quo{lpsolve} ILP backend in experiments,
however Gurobi cannot improve the situation much
because most instances timed out during instantiation.
From the authors of \henry\ and \phillip\ we obtained
\rev{the rewriting}{a transformed version }of \accel\ that was used to produce
their published results;
unfortunately that rewriting is incompatible
with the current version of \henry\ as well as \phillip,
however we noticed that the rewriting
makes some simplifying assumptions on \accel:
it interprets constants in goals as sort names,
i.e., distinct entities in the goal can never denote the same entity
(which makes, e.g., coreference resolution impossible);
moreover sort names are compiled into predicates
which creates many distinct predicates of lower arity
and makes instantiation easier.
Also assumption constraints are not realized.
For \rev{the more general case that our encodings can solve,}{reasoning with the complete set of rules in \accel,}
our approach is significantly faster than
the state-of-the-art solver \phillip\rev{\ on the \accel\ benchmark}{}.

\mytableGeneralVsPython{tbp}
\paragraph{Skolemization.}
Table~\ref{tblASPPythonSkolem} compares the encodings
and objectives from Table~\ref{tblGeneral}
with their counterparts using Python Skolemization.
For a \rev{better}{fair} comparison,
we here do not compare with pure \gringo+\clasp,
but we use our algorithm in Python in both cases
\rev{}{(even for ASP encodings that do not use Python Skolemization)}:
this explains differences to Table~\ref{tblGeneral}%
\rev{}{\ and why we do not experiment with \wasp\ here}.
For these and previous experiments we use only acyclic axioms
and do not limit Skolemization, so the only difference
between the program is the shape of constant symbols:
nested uninterpreted function terms of form $s(\cdot,\cdot,\cdot)$
versus constants of form $p_i$, $i \ins \mb{N}$,
generated with Python.
Although Python Skolemization provides higher flexibility
than uninterpreted function terms,
there is no noticeable effect on efficiency
of the \rev{20}{17} easiest instances,
and across all 50 instances,
Python Skolemization has a positive effect
on efficiency of the
\rev{\bwda\  and \fwda\ encodings,
but a small negative effect on \bwdai.}%
{\bwdai\  and \fwda\ encodings,
while performance of the most efficient encoding
\bwda\ is unchanged.}%

\rev{From these measurements it}{It} is not apparent
why \rev{\fwda\ shows better}{
some encodings improve} performance
with Python Skolemization.
\rev{We know, that the main difference
between \fwda\ and \bwd\ encodings
is the structure of the encoding.}{}%
\rev{In theory}{Structurally},
the instantiated program entering the solver
is exactly the same in both Skolemization methods,
except for Skolem constants
\rev{}{instead of uninterpreted function terms}
in the symbol table.
However, we \rev{}{additionally }observed,
that the order of rules \rev{which are }{}created by \gringo\rev{,}{}
changes between both Skolemization methods.
From this we conclude that the order of rules matters,
and that there is potential for optimization in solvers,
moreover this suggests that \clasp\ is sensitive
to the order of rules it consumes before solving.
(We \rev{here }{}conjecture, that efficiency is not affected
by the \rev{}{form of strings in the }symbol table\rev{ which is used for printing the result}{}.)

We conclude that
\rev{using Python Skolemization
is not a big disadvantage in terms of performance,
so obtaining the additional}{in addition to the}
flexibility \rev{}{of Python Skolemization}
\rev{does not require us to sacrifice }%
{we can gain} efficiency.

\mytableCyclicityEquivalenceOnDemand{tbp}
\paragraph{On-Demand Constraints.}
Table~\ref{tblASPvsOnDemand} shows experimental results
for comparing two methods for ensuring
acyclicity of the relation $\mibelow$
and transitivity of the relation $\mieq$.
The \rev{results marked as}{rows with (R)} use encodings as given in Section~\ref{secMainEncodings}
while those \rev{marked}{with} (O)
use %
on-demand constraints %
as described in Section~\ref{secOnDemand}.
\rev{}{For a fair comparison, all runs were performed
with algorithms based on the \clasp\ Python API (Section~\ref{secImpl}).}

We observe that on-demand constraints significantly
reduce instantiation time and memory usage
in all encodings and all objective functions.
\rev{%
Although in some cases both memory usage and time usage decrease on average,
the number of instances solved optimally actually decreases for \accelobj.
This seems counterintuitive,
however it is not an inconsistency:
on-demand constraints reduce the time spent in grounding significantly,
whereas the time spent in solving can increase or decrease,
depending on the number of on-demand constraints that is generated
until finding the optimal solution.
As an example, instance \#13 of the \accel\ benchmark
requires $\myodc=562$K on-demand constraints and $\myconf=505$K conflicts
for finding the first optimal answer set;
while the pure ASP solution requires just $\myconf=58$K conflicts,
moreover grounding time \mygrd\ decreases from $55$ sec to $5$ sec,
solving time \myslv\ increases from $60$ sec to $200$ sec.
Timeouts are counted as the maximum time,
so solving several instances significantly faster
can make up for solving fewer instances optimally.
From 50 \accel\ instances,
5 (none of them among the 20 easiest) show a slowdown,
the others obtain an overall gain in efficiency.

}{}%
We can see the difference between instantiating
the full encodings or encodings without rules
\eqref{eqEqTransitive} and \eqref{eqBelow2} in column $\mygrd$.
We observe that instantiating transitivity and acyclicity
dominates the instantiation time,
and that back-chaining in ASP is fast.
Therefore we did not perform experiments
for creating the proof graph outside ASP
(this would correspond to the architecture of \phillip,
where the proof graph is created in C++ and solved in ILP).

\rev{}{Interestingly, for the \accelobj\ objective, encoding \bwdai\ outperforms \bwda\ with on-demand constraints (although by a small amount, and although the instantiation time of the easiest 17 instances of \bwda\ is smaller than the one of \bwdai).}

\rev{In summary, it depends on the instance whether on-demand constraints
provide a significant gain or loss in performance,
although they always yield}{%
As reported in the ILP-based solvers \henry\ and \phillip,
on-demand constraints turn out to be important for managing
bigger instances in this reasoning problem:
we observe increased performance and}
a significant reduction in memory usage.

\mytableLimitedSkolemization{tbp}
\paragraph{Cyclic Theories and Limited Skolemization.}
Table~\ref{tblLimitedSkolemization} shows the results of experiments
with cyclic theories and limited value invention
as defined in Section~\ref{secFlexibleValueInvention}.
For encoding \bwda\ and all objective functions,
we evaluate
the acyclic theory with unlimited $\extfun{sk^{\infty}}$ Skolemization,
and the theory including the cyclic axiom
with parent- and rule-based Skolemization limits
$\extfun{sk^\alpha}$ with
$\alpha \ins \{ P^1, P^2, G^1, G^2\}$.

We show only \accelobj\ and \waobj\ \rev{}{with \bwda,} because
\cardobj\rev{\ and \accelobj}{\ and other encodings}
show the same \rev{effects.
Moreover we only show encoding \bwda\ because
other encodings show the same}{}trends.

We observe that
\rev{}{time (\myt)} and memory (\mym) \rev{requirement}{usage,}
number of generated Skolem constants (\myskc),
grounding time (\mygrd), and
size of the instantiation (\myrul),
are ordered $P^1 \lts P^2 \lts \infty \lts G^1 \lts G^2$
for \rev{all encodings and }{both} objective functions.
Solving time \rev{}{(\myslv),
choices ($\mychc$), and conflicts ($\myconf$),
are also nearly always} ordered like that.
\rev{with the only exception of \bwda\ encoding with \waobj\ objective function,
where $\infty$ finds the optimal solution faster than $P^2$.
We can explain this by looking into solver metrics not shown in Table~\ref{tblLimitedSkolemization}:
in configuration \bwda/\waobj,
the $P^2$ Skolem limitation produces the highest number of conflicts
in the search, even higher than $G^2$, although $P^2$ has
significantly smaller instantiation and number of choices.
While $\infty$ has a higher number of choices, it has a lower number of conflicts and finds optimal solutions faster.
The reason for this could be,
that $P^2$ is less constrained than $\infty$
and therefore produces symmetries which
require a higher amount of conflicts
for proving optimality of the solution.}{}

Regarding the objective function,
\rev{limiting value invention ($P^1$) leads to higher costs of the optimal solution wrt.\ \accelobj,
because having longer inference paths with deeper value invention allows higher coherence. %
Weighted abduction (\waobj) shows a very clear difference in the cost of optimal solutions:
limiting value invention increases cost of the optimal solution,
and this increase is directly connected with the amount \myskc\ of available invented values.}%
{the $\infty$ method does not use the (single) cyclic axiom in \accel,
while the other methods use that axiom.
$P^1$ and $P^2$ permit a limited amount of value invention
based on invented values, and allow fewer inferences than $\infty$
on the acyclic axioms,
which results in a higher cost of the optimal solution.
$G_1$ and $G_2$ block value invention only when it reaches the same rule
(not other rules),
therefore they allow a superset of the inferences of $\infty$.
This explains, why \myobj\ of $\infty$
is above the one of $P_2$ and below the one of $G_1$
(recall that we display \myobj\ only for the \rev{20 easiest}{17} instances
where all runs found the optimum).}

Regarding efficiency,
\rev{we see that }{one or }two generations of invented values across rules (\rev{}{$G^1$, }$G^2$)
\rev{already cause}{drastically increase} memory exhaustion\rev{ for several instances,
and even allowing a single generation of invented values for each rule ($G^1$)
exceeds the timeout in several cases.}%
{, while strictly limiting value invention ($P^1$)
makes the problem easy to solve and impairs solution quality.}

\mytableAssumptionUniqueConstraints{tbp}
\paragraph{Global Constraints.}
Table~\ref{tblAssumptionUniqueConstraints}
shows a comparison between realizing unique-slot constraints
and assumption constraints in ASP constraints
versus not considering these constraints.

\rev{Global constraints slightly increase the memory requirement.
Interestingly, presence of constraints makes it easier
to find the optimal solution
for all encodings and objective functions.}%
{Global constraints have no significant effect on efficiency. }

We see no effect on the objective function
when removing global constraints.
This seems counterintuitive:
more constraints should intuitively
increase cost of the optimal solution.
It turns out that these cases are rare:
we can observe an increased cost
of optimal solutions
if we limit Skolemization using method $P^1$%
\rev{:
then the average \waobj\ objective over all instances
is 671.2 with global constraints,
while it is 667.9 without them}{}.
We conclude that global constraints \rev{}{in the \accel\ benchmark }have
a small impact on solution quality.

\paragraph{Other Experiments.}
For combining on-demand constraints with optimization,
we also investigated alternatives to
\rev{Algorithm~\ref{algoFindOptimalModel}}{algorithm \textsc{FindOptimalModel}
(see Section~\ref{secImpl} and the appendix)}
where we used \clasp\ assumptions and \clasp\ externals
for deactivating certain optimization criteria
during some parts of the search.
These \rev{methods}{alternatives} perform significantly worse%
\rev{ than Algorithm~\ref{algoFindOptimalModel}}{},
moreover they are involved and require
rewriting weak constraints into normal rules,
therefore we decided to omit
further details about these experiments.

\rev{}{Realizing global constraints of \accel\ also in an on-demand manner
did not yield significantly different results from using the pure ASP versions,
therefore we omit these results from the presentation.}

We experimented with projecting answer sets
to the atoms that are relevant for the objective function,
which yielded a significant reduction in log file size
(because we print the solution)
but no significant reduction in time or memory.

\section{Related Work}
\label{secRelated}

The idea of abduction goes back to Peirce \cite{Peirce1955}
and was later formalized in logic.

Abductive Logic Programming (ALP)
is an extension of logic programs
with abduction and integrity constraints.
Kakas et al.\ \cite{Kakas1992}
discuss ALP and applications,
in particular they relate Answer Set Programming and abduction.
Fung et al.\ describe the IFF proof procedure \cite{Fung1997}
which is a FOL rewriting that is sound and complete
for performing abduction in a fragment of ALP
with only classical negation and specific safety constraints.
Denecker et al.\ \cite{Denecker1998}
describe SLDNFA-resolution
which is an extension of SLDNF resolution
for performing abduction in ALP
in the presence of negation as failure.
They describe a way to
\quo{avoid Skolemization by variable renaming}
which is exactly what we found to increase performance
in flexible Skolemization
(recall that we create numbered constants $p_i$
instead of structured terms $s(\cdots)$ in Python).
Kakas et al.\ describe the $\mathcal{A}$-System
for evaluating ALP using an algorithm that interleaves
instantiation of variables and constraint solving
\cite{Kakas2001}.
The CIFF framework \cite{Mancarella2009}
is conceptually similar to the $\mathcal{A}$-System
but it allows a more relaxed use of negation.
The \sciff\ framework \cite{Alberti2008}
relaxes some restrictions of CIFF
and provides facilities for modeling agent interactions.
In \cite{Gavanelli2015}, \sciff\ was used to realize semantics of
Datalog$^{\pm}$ \cite{Cali2009} which natively supports existentials
in rule heads (i.e., value invention) as opposed to ASP.
The focus of \sciff\ is on finding abductive explanations,
while our focus is to find preferred abductive explanations
according to objective functions.
Realizing objective functions requires modifying the \sciff\ engine
(Evelina Lamma 2015, personal communication)
therefore we did not perform experiments comparing \sciff\ with our encodings.

Implementations of ALP, have in common that they are
based on evaluation strategies
similar to Prolog \cite{Mancarella2009}.
In \cite{Mancarella2009},
CIFF is compared with ASP on the example of n-queens
and the authors emphasize that CIFF has more power
due to its partial non-ground evaluation.
However, they use a non-optimized n-queens encoding for that comparison,
and optimized n-queens encodings for \clingo \cite{Gebser2012aspbook}
are known to yield orders of magnitude better performance than naive encodings,
hence partial non-ground evaluation is not necessarily a guarantee for better performance.
Different from CIFF and earlier ALP implementations,
our approach \rev{is to first }{}instantiate\rev{}{s
one Boolean variable for each node in }the \rev{whole }{}potential proof graph
and then search\rev{}{es }for the best solution%
\rev{ (note that, even with on-demand constraints,
we instantiate the whole potential proof graph)}%
{, while methods that create nodes on demand (such as CIFF)
can completely eliminate certain nodes from instantiation,
while instantiating other nodes multiple times}.

The AAA (ATMS-based Abduction Algorithm) reasoner
\cite{Ng1992thesis,Ng1992kr} \rev{}{combines Prolog resolution
with ATMS-based caching for realizing abduction.}
For \accel, AAA realizes \cardobj\ and \accelobj\ metrics
and \rev{realizes}{enforces }assumption-nogoods and
unique-slot constraints in dedicated \rev{}{(imperative) }procedures%
\rev{that also consider equivalence between terms}{}.

The \henry\ reasoner \cite{Inoue2013}
realizes \waobj\ by creating an ILP instance with C++
using back-chaining
and then finding optimal solutions for the ILP instance.
The newest version of \henry\ is called \phillip\ \cite{Yamamoto2015};
this solver adds heuristics that partially instantiate the proof tree
according to relatedness between predicates
(although without formal proof of the correctness
or worst-case approximation error).
This two-step approach is similar to our approach in ASP:
our encodings cause the ASP grounder to perform
back-chaining in the knowledge base,
and after instantiation the solver searches for
optimal solutions satisfying all rules and constraints.
A big performance improvement for \henry\
was the usage of Cutting Plane Inference \cite{Inoue2013}.
We mimic the approach with on-demand constraints in ASP,
and we can observe similar improvements in instantiation size and time,
however solve time increases by a larger amount for many instances,
hence this approach is not sufficient for achieving a
corresponding performance boost in ASP.
Within the ASP community,
on-demand constraints are related to methods of
lazy instantiation of ASP programs,
as done in the solvers
ASPeRiX \cite{Lefevre2015},
OMiGA \cite{DaoTran2012},
Galliwasp \cite{Marple2013galliwasp},
and recently also in the IDP system \cite{DeCat2015}.
These systems apply lazy instantiation to all rules and constraints in the program,
whereas we make only certain problematic constraints lazy.

\emph{Probabilistic} abduction was realized in Markov Logic %
\cite{Richardson2006}
in the Alchemy system \cite{Kok2010}
although without value invention \cite{Blythe2011,Singla2011},
i.e., existential variables in rule heads are naively instantiated with all ground terms in the program.
A corresponding ASP encoding for the non-probabilistic case for \cardobj\ exists \cite[\cbfan]{Schuller2015rcra},
however it shows prohibitively bad performance.

The termination proofs we do
are related to the notion of \emph{Liberal Safety} in \hex\ programs
\cite{Eiter2013liberalsafety},
however Liberal Safety requires either specific
acyclicity conditions (which are absent in our encodings),
or conditions on finiteness of the domain of
certain attributes of the external atom
(that our Skolemization atoms do not fulfill).
Hence we had to prove termination without using Liberal Safety.

\rev{}{In the area of Automated Theorem Proving,
algorithms search for finite models (or theorems, unsatisfiability proofs)
in full first order logic without enforcing UNA and including native support for Skolemization
(cf.\ \cite{Sutcliffe2009tptp}).
These algorithms focus on finding a feasible solution
and do not contain support for preferences (optimization criteria).
However, the main emphasis of our abduction problems is to find solutions with optimal cost
(recall that our problems always have the trivial solution to abduce all input atoms).
To tackle our abduction problem with such theorem provers,
it would be necessary to transform the optimization problem
into a decision problem and perform a search over the optimization criterion,
calling the prover several times.
Related to theorem proving,
a hypertableaux algorithm for coreference resolution
is described in \cite{Baumgartner2000}.
This algorithm is inspired by weighted abduction,
however it does not use preferences and relies solely on inconsistency for eliminating undesired solutions.}

\paragraph{Computational Complexity.}
The complexity of abduction in \emph{propositional} theories
in the presence of the \cardobj\ objective
has been analyzed in \cite{Bylander1991,Eiter1994},
and in \cite{Eiter1997abductionlpcomplexity},
the propositional case of abduction in logic programs
is studied and extended to
function-free logic programming abduction (Sec.~6),
under the restriction that only constants from
observations and knowledge base
(there called \quo{manifestations} and \quo{program})
are used and that the UNA holds for all terms.
However, in our variant of abduction
the optimal solution may use a set (of unspecified size)
of constants that are not present in the input
and there is potential equality
among certain input constants and constants originating in value invention.
Hence, existing results can be seen as lower bounds for hardness
but do not directly carry over to our scenario.

In an acyclic theory,
our reasoning problem is related to non-recursive
negation-free Datalog theories and
non-recursive logic programming with equality,
which has been studied
(although not with respect to abductive reasoning)
in \cite{Dantsin2001}.

Creating the largest possible proof graph
for a given goal and knowledge base
can be done by reversing axioms
and evaluating them with the goal;
the complexity of this problem
(definite not range-restricted
logic program without function symbols)
was shown to be PSPACE-complete
\cite[Thm.~4.1]{Vorobyov1998}.

\section{Conclusion}
\label{secConclusion}

We have created a flexible and publicly available
framework for realizing variations of cost-based FO Horn abduction
represented in the declarative reasoning framework of Answer Set Programming%
\rev{}{~\cite{Lifschitz2008}}
that allows us
(i) to modularly encode additional objective functions based on the abductive proof graph, and
(ii) to add global constraints of arbitrary complexity and size.
\rev{}{Our encodings use a modular translation of axioms into ASP rules,
i.e., each axiom can be translated independent from other axioms.}
As preference relations we realized
cardinality-minimality, coherence \cite{Ng1992kr},
and weighted abduction \cite{Hobbs1993,Stickel1989}.
We evaluated our framework on the \accel\ benchmark\rev{}{~\cite{Ng1992kr}}
and found that we have significantly higher performance
than state-of-the-art solver \phillip\ \cite{Yamamoto2015}
which is the successor system of \henry \cite{Inoue2013}.
In our experiments, \wasp\ \rev{}{\cite{Alviano2015wasp}}
solves instances faster than \clasp\rev{}{~\cite{Gebser2015clasp3}},
however \wasp\ uses more memory for programs with a
high degree of nondeterminism.

For realizing value invention we experimented with uninterpreted functions
and with external computations providing new values to the program.
Performing Skolemization with external computations
provides fine-grained control for deciding when to instantiate a term
and when to refuse further instantiation.
This allows us to ensure and formally prove decidability
when performing abduction in cyclic FO Horn knowledge bases.
Fortunately, this flexibility does not impair computational efficiency.

An important topic in this research is encoding the proof graph.
Usually, in ASP we are not interested in the order of inferences made in the program
or in the dependencies or equivalences between atoms --- those are handled transparently in the solver.
However, for modeling preference functions \accelobj\ and \waobj,
which are defined on proof graphs,
we must explicitly represent a proof graph,
including back-chaining and unification, in our ASP encoding.
\rev{For reasons of comparison we}{We} also experiment with an
\rev{}{alternative encoding (\fwda) for representing objective \cardobj: this}
encoding \rev{that }{}performs abduction
without representing a proof graph\rev{ (\fwda):
this encoding is only suitable for representing objective \cardobj,
solves a different reasoning problem than the other encodings
(it performs}{, it is based on }forward inference\rev{s)}{,
has good performance with the \wasp\ solver}.

\rev{}{The \bwda\ encoding performs best,
intuitively because it makes the most strict canonicalization
operations on the graph by requiring factoring to happen only
with abduced atoms.
Proof cost is not affected by this canonicalization,
however proof graphs will contain duplicate inferences
for atoms that are equivalent due to term equivalence:
these atoms could be first factored and inferred later.
Yet this seemingly wasteful proof graph does not diminish
performance in ASP,
because we anyway need to instantiate the whole
potential proof graph,
so all inferences trees are available
even if we do permit factoring anywhere in the tree.
Postponing factoring to abduced atoms
has the effect that we need to handle term equivalence
only for these atoms,
which is an advantage for efficiency.}

As ASP provides no native support for term equivalence,
we \rev{also need to }{}encode equivalence and unification explicitly.
We guess an equivalence relation
and check its reflexivity, symmetry, transitivity with constraints.
Explicit representation of transitivity of equivalence has been shown to be a major
performance issue in \henry, as it causes instantiation of a cubic amount of rules \cite{Inoue2013}.
This performance issue also becomes apparent in our ASP representation,
and we apply the solution from \henry\ to our encodings by using on-demand constraints,
which we describe formally in the \hex\ formalism \cite{Eiter2015hex}.
Realizing on-demand constraints in presence of optimization is nontrivial in current solvers,
and we describe an algorithm based on the Python API
of \rev{the \clasp+\gringo\ solver family }{\clingo }%
\cite{Gebser2014clingorepext}.
\rev{On-demand constraints do not consistently improve solving time in our ASP encodings,
as opposed to the results reported for \henry\ for weighted abduction \cite{Inoue2013}:
for the simple objective function \cardobj\ we obtain a significant speedup and memory reduction
for all encodings we study,
however for more complex objective functions (\accelobj, \waobj) we achieve memory reduction
but solving times sometimes increase drastically
and a large amount of on-demand constraints is generated.
Hence, depending on the instance and objective function,
on-demand constraints can be used to trade time for space,
or in some cases achieve overall time and space reduction.
}%
{On-demand constraints significantly reduce memory usage
by partially instantiating transitivity constraints
of the term equivalence relation
(note that the potential proof graph is still fully instantiated).
This is consistent with results reported for \henry\ and \phillip\ for
weighted abduction \cite{Inoue2013} with ILP as a solver backend
and not surprising as ASP and ILP are related methods for solving combinatorial problems \cite{Liu2012}.}

\paragraph{Future work.}
The major motivation for this work was to obtain a more flexible framework
where variations of objective functions and constraints on abduction can be studied.
In the future we intend to perform research in this direction.
Moreover we want to apply our encodings to other datasets
like the one derived from the Recognizing Textual Entailment (RTE) \cite{Dagan2006} challenge%
\rev{ which will bring additional challenges due to its larger scale.
We expect, that on-demand constraints will be important
for handling space consumption in this dataset
similar as in \henry\ where on-demand constraints were added
to tackle RTE instances \cite{Inoue2013}.}{.}

Among the encodings we experiment with,
the most obvious and straightforward encoding (\bwdg)
has the worst performance,
and small encoding changes as well as
bigger changes, that realize symmetry breaking
based on a theoretical analysis of the problem,
are required for achieve acceptable performance (\bwda).
\rev{}{Interestingly, the \wasp\ solver is able to
compensate for the more nondeterministic representation:
it performs similar on \bwdg\ and \bwdai\ encodings,
opposed to \clasp\ which performs significantly worse on \bwdg.}
We conclude that automatic program optimization
and supporting tools for diagnosing performance issues
are open problems in ASP and fruitful topics for future work.
\rev{Regarding computational efficiency,
a major performance issue is transitivity of the equivalence relation.
In \henry\ and \phillip, this problem has been solved
by adding these constraints on-demand when they are violated \cite{Inoue2013}%
and we attempt to solve it analogously in ASP.
However this did not yield a similar performance improvement.
This raises new research questions for the ASP community,
because ASP and ILP are related formalisms \cite{Liu2012}
and we see no apparent theoretical limitation for handling this problem in ASP,
nor for preventing the technique used in ILP to be successful also in ASP.}{}

\section*{Acknowledgements}

We thank Naoya Inoue, Evelina Lamma, Christoph Redl,
and the \rev{RCRA}{anonymous }reviewers and RCRA workshop participants
for constructive feedback about this work.
We thank Mario Alviano, Carmine Dodaro,
Roland Kaminski, and Benjamin Kaufmann for support regarding
\clasp, \gringo, and \wasp.

\bibliographystyle{alpha}

\newcommand{\etalchar}[1]{$^{#1}$}

\ifappendix
\clearpage
\allowdisplaybreaks
\section{Appendix: Complete Encodings}

\rev{}{We give full encodings in the following.}

\paragraph{Complete Encoding \bwd.}
Each axiom of form \eqref{eqAx} is rewritten into
the following set of ASP rules:
\begin{eqnarray*}
  &&\mimayinfervia(a,c(q,Y_1,\ldots,Y_m),l(Z_1,\ldots,Z_v)) \lars
      \notag \\
  &&\rulebodyhspace
      \mipot(c(q,Y_1,\ldots,Y_m)),\
      Z_1 \eqs s(a,1,Y_1,\ldots,Y_m),\
      \ldots, Z_v \eqs s(a,v,Y_1,\ldots,Y_m)\quad \\
  &&\miinferenceneeds(c(q,Y_1,\ldots,Y_m),a,c(p_i,X^i_1,\ldots,X^i_{k_i})) \lars \\
  &&\rulebodyhspace \mimayinfervia(a,c(q,Y_1,\ldots,Y_m),l(Z_1,\ldots,Z_v))
  \qquad\qquad \text{for } i \ins \{1,\ldots,r\} \notag
\end{eqnarray*}
where $a$ is a unique identifier for that particular axiom
and $Z_1,\ldots,Z_v \eqs \cX \setminuss \cY$.

The encoding contains the following rules.
\begin{align*}
  \mipot(X) &\lars \migoal(X). \\
  \mipot(P) &\lars \miinferenceneeds(\_,\_,P) \\
  \mitrue(P) &\lars \migoal(P). \\
  1 \les &\{\ \miinfer(P)\ ;\ \mifai(P)\ \} \les 1 \lars \mitrue(P).\\
  1 \les &\{\ \miinfervia(R,P) : \mimayinfervia(R,P,\_)\ \} \les 1 \lars \miinfer(P). \\
  \mitrue(Q) &\lars \miinfervia(R,P), \miinferenceneeds(P,R,Q). \\
  hu(X) &\lars \mipot(c(\_,X,\_)). \\
  hu(X) &\lars \mipot(c(\_,\_,X)). \\
  \miuhu(X) &\lars hu(X), \naf\ \misortname(X). \\
  &\{\ eq(A,B) : \miuhu(A),\ \miuhu(B),\ A \neqs B\ \} \lars. \\
  eq(A,A) &\lars hu(A). \\
  &\lars eq(A,B), \naf\ eq(B,A). \\
  &\lars eq(A,B), eq(B,C), A \rev{\lts}{\neqs} B, B \rev{\lts}{\neqs} C, \rev{}{A \neqs C,} \naf\ eq(A,C).
\end{align*}

\paragraph{Complete Factoring Encoding \bwdg.}

\begin{align*}
  \mibelow(P,Q) &\lars \miinfervia(R,P), \miinferenceneeds(P,R,Q). \\
  \mibelow(P,Q) &\lars \mifactorvia(P,Q). \\
  \mibelow(A,C) &\lars \mibelow(A,B), \mibelow(B,C). \\
  1 \les \{\ \mifactor(P)\ ;\ \miabduce(P)\ \} \les 1 &\lars \mifai(P). \\
  \mifactorvia(c(P,S_1,O_1),c(P,S_2,O_2)) &\lars
    \mifactor(c(P,S_1,O_1)), \miinfer(c(P,S_2,O_2)), eq(S_1,S_2), \notag\\
  &\quad eq(O_1,O_2), \naf\ \mibelow(c(P,S_2,O_2),c(P,S_1,O_1)). \\
  \mifactorvia(c(P,S_1,O_1),c(P,S_2,O_2)) &\lars \notag
    \mifactor(c(P,S_1,O_1)), \miabduce(c(P,S_2,O_2)), \notag \\
  &\quad eq(S_1,S_2), eq(O_1,O_2). \\
  \mifactorOk(P) &\lars \mifactorvia(P,\_). \\
  &\lars \mifactor(P), \naf\ \mifactorOk(P).
\end{align*}

\paragraph{Complete Factoring Encoding \bwdai.}

\begin{align*}
  \mibelow(P,Q) &\lars \miinfervia(R,P), \miinferenceneeds(P,R,Q). \\
  \mibelow(P,Q) &\lars \mifactorvia(P,Q). \\
  \mibelow(A,C) &\lars \mibelow(A,B), \mibelow(B,C). \\
  \mifactorviai(c(P,S_1,O_1),c(P,S_2,O_2)) &\lars
    \mifai(c(P,S_1,O_1)), \miinfer(c(P,S_2,O_2)), eq(S_1,S_2), \notag\\
  &\quad eq(O_1,O_2), \naf\ \mibelow(c(P,S_2,O_2),c(P,S_1,O_1)). \\
  \mifactori(P) &\lars \mifactorviai(P,\_). \\
  \mifa(P) &\lars \mifai(P), \naf\ \mifactori(P). \\
  \mifactorvia(A,B) &\lars \mifactorviai(A,B). \\
  \mifactorcluster(c(P,S_2,O_2),c(P,S_1,O_1)) &\lars \mifa(c(P,S_1,O_1)), \mifa(c(P,S_2,O_2)), eq(S_1,S_2), \notag \\
    &\quad eq(O_1,O_2), c(P,S_1,O_1) \lts c(P,S_2,O_2). \\
  \mifactorclusterabove(A) &\lars \mifactorcluster(A,\_). \\
  \mifactorvia(A,B) &\lars \mifactorcluster(A,B), \notag \\
    &\quad \naf\ \mifactorclusterabove(B).  \\
  \mifactor(P) &\lars \mifactorvia(P,\_). \\
  \miabduce(P) &\lars \mifa(P), \naf\ \mifactor(P).
\end{align*}

\paragraph{Complete Factoring Encoding \bwda.}

\begin{align*}
  \mifa(P) &\lars \mifai(P). \\
  \mifactorcluster(c(P,S_2,O_2),c(P,S_1,O_1)) &\lars \mifa(c(P,S_1,O_1)), \mifa(c(P,S_2,O_2)), eq(S_1,S_2), \notag \\
    &\quad eq(O_1,O_2), c(P,S_1,O_1) \lts c(P,S_2,O_2). \\
  \mifactorclusterabove(A) &\lars \mifactorcluster(A,\_). \\
  \mifactorvia(A,B) &\lars \mifactorcluster(A,B), \notag \\
    &\quad \naf\ \mifactorclusterabove(B).  \\
  \mifactor(P) &\lars \mifactorvia(P,\_). \\
  \miabduce(P) &\lars \mifa(P), \naf\ \mifactor(P).
\end{align*}

\paragraph{Complete Encoding \fwda.}

Each axiom of form \eqref{eqAx} is rewritten into
the following set of ASP rules:
\begin{align*}
  \miinfer(c(q,Y_1,\ldots,Y_m)) &\lars
    \mitrue(c(p_1,X^1_1,\ldots,X^1_{k_1})),
    \ldots,
    \mitrue(c(p_r,X^r_1,\ldots,X^r_{k_r})). \\
  \mipot(c(p_i,X^i_1,\ldots,X^i_{k_1})) &\lars
      Z_1 \eqs s(a,1,Y_1,\ldots,Y_m), \ldots,
       Z_v \eqs s(a,v,Y_1,\ldots,Y_m), \\
  &\quad \mipot(c(q,Y_1,\ldots,Y_m)).
         \qquad\qquad\qquad\qquad\qquad\qquad \text{ for } i \ins \{ 1, \ldots, r \}
\end{align*}
where $a$ is a unique identifier for that particular axiom
and $Z_1,\ldots,Z_v \eqs \cX \setminuss \cY$.

The encoding \fwda\ then contains the following rules.
\begin{align*}
  \mipot(X) &\lars \migoal(X). \\
  \{\ \mifai(X) : \mipot(X)\ \} &\lars. \\
  \mitrue(X) &\lars \mifai(X). \\
  \mitrue(X) &\lars \miinfer(X). \\
  &\lars \migoal(A), \naf\ \mitrue(A). \\
  \mifa(X) &\lars \mifai(X), \naf\ \miinfer(X). \\
  hu(X) &\lars \mipot(c(\_,X,\_)). \\
  hu(X) &\lars \mipot(c(\_,\_,X)). \\
  \miuhu(X) &\lars hu(X), \naf\ \misortname(X). \\
  \{\ eq(A,B) : \miuhu(A),\ \miuhu(B),\ A \neqs B\ \} &\lars. \\
  eq(A,A) &\lars hu(A). \\
  &\lars eq(A,B), \naf\ eq(B,A). \\
  &\lars eq(A,B), eq(B,C), A \rev{\lts}{{\neq}} B, B \rev{\lts}{{\neq}} C, \rev{}{A {\neq} C,} \naf\ eq(A,C). \\
  \mifactorcluster(c(P,S_2,O_2),c(P,S_1,O_1)) &\lars \mifa(c(P,S_1,O_1)), \mifa(c(P,S_2,O_2)), eq(S_1,S_2), \notag \\
    &\quad eq(O_1,O_2), c(P,S_1,O_1) \lts c(P,S_2,O_2). \\
  \mifactorclusterabove(A) &\lars \mifactorcluster(A,\_). \\
  \mifactorvia(A,B) &\lars \mifactorcluster(A,B), \notag \\
    &\quad \naf\ \mifactorclusterabove(B).  \\
  \mifactor(P) &\lars \mifactorvia(P,\_). \\
  \miabduce(P) &\lars \mifa(P), \naf\ \mifactor(P).
\end{align*}

\rev{}{
\section{Appendix: Running Example ASP Encoding and Answer Set}

We next give the ASP input instance for our running example
(Example~\ref{mainexample}) when used with \bwd\ encodings.
We then give representative parts of an answer set
describing the abductive explanation shown in Figure~\ref{figExWA}.

\subsection{Rewriting of Axioms, Goal, and Sortnames}

As described in Section~\ref{secMainEncodings},
given the an abduction instance $A \eqs (B,O,S)$
of Example~\ref{mainexample} we create the following ASP code.

We represent the goal in terms of facts \eqref{eqGoalFact}.
\begin{align*}
  &\migoal(c(\miname,m,\mimary)).\quad 
  &&\migoal(c(\milost,m,f)). \quad
  &&\migoal(c(\mifatherof,f,m)). \\
  &\migoal(c(\miinst,s,\mifemale)).\quad 
  &&\migoal(c(\miis,s,\midepressed)).
\end{align*}
We represent sort names as follows.
\begin{align*}
  &\misortname(\midepressed).\quad
  &&\misortname(\midead).\quad
  &&\misortname(\miperson). \\
  &\misortname(\mifemale).\quad
  &&\misortname(\mimale).
\end{align*}
According to the rewriting \eqref{eqFormalBwd}
in Section~\ref{secBwd},
we rewrite axioms of Example~\ref{mainexample} as follows.

\newcommand{\eqhspacer}{\hspace*{15em}}
Axiom \eqref{eqEx1} \quo{$\miinst(X,\mimale) \lax \mifatherof(X,Y)$}
is rewritten into the following ASP rules.
\begin{align*}
  &\mimayinfervia(r_4,c(\miinst,X,\mimale),l(Y)) \lars 
   \mipot(c(\miinst,X,\mimale)), Y \eqs s(r_4,``Y",X). \\
  &\miinferenceneeds(c(\miinst,X,\mimale),r_4,c(\mifatherof,X,Y)) \lars \\
   &\eqhspacer \mimayinfervia(r_4,c(\miinst,X,\mimale),l(Y)). \\
  &\minumberofbodies(r_4,1).
\end{align*}
Note the Skolemization of variable $Y$ which exists
only in the body of \eqref{eqEx1} but not in the head.
Also note that we use rule identifiers that are synchronized
with the original axiom numbers, i.e., for \eqref{eqEx1} we use $r_4$.

Axiom \eqref{eqEx2} \quo{$\miinst(X,\mifemale) \lax \miname(X,\mimary)$}
is rewritten into the following ASP rules.
\begin{align*}
  &\mimayinfervia(r_5,c(\miinst,X,\mifemale),l) \lars
    \mipot(c(\miinst,X,\mifemale)). \\
  &\miinferenceneeds(c(\miinst,X,\mifemale),r_5,c(\miname,X,\mimary)) \lars \\
   &\eqhspacer \mimayinfervia(r_5,c(\miinst,X,\mifemale),l). \\
  &\minumberofbodies(r_5,1).
\end{align*}
Note that in this axiom there is no Skolemization,
so $l$ has no arguments (however we still need it to keep the arity of
$\mimayinfervia$ the same throughout the encoding.

Axiom \eqref{eqEx5} \quo{$\miimportantfor(Y,X) \lax \mifatherof(Y,X)$}
is rewritten into the following ASP rules.
\begin{align*}
  &\mimayinfervia(r_6,c(\miimportantfor,Y,X),l) \lars
    \mipot(c(\miimportantfor,Y,X)). \\
  &\miinferenceneeds(c(\miimportantfor,Y,X),r_6,c(\mifatherof,Y,X)) \lars \\
   &\eqhspacer \mimayinfervia(r_6,c(\miimportantfor,Y,X),l). \\
  &\minumberofbodies(r_6,1).
\end{align*}
Axiom \eqref{eqEx7} \quo{$\miinst(X,\miperson) \lax \miinst(X,\mimale)$}
is rewritten into the following ASP rules.
\begin{align*}
  &\mimayinfervia(r_7,c(\miinst,X,\miperson),l) \lars
    \mipot(c(\miinst,X,\miperson)). \\
  &\miinferenceneeds(c(\miinst,X,\miperson),r_7,c(\miinst,X,\mimale)) \lars 
   \mimayinfervia(r_7,c(\miinst,X,\miperson),l). \\
  &\minumberofbodies(r_7,1).
\end{align*}
Axiom \eqref{eqEx3} \quo{$\miis(X,\midepressed) \lax \miinst(X,\mipessimist)$}
is rewritten into the following ASP rules.
\begin{align*}
  &\mimayinfervia(r_8,c(\miis,X,\midepressed),l) \lars
    \mipot(c(\miis,X,\midepressed)). \\
  &\miinferenceneeds(c(\miis,X,\midepressed),r_8,c(\miinst,X,\mipessimist)) \lars \\
   &\eqhspacer \mimayinfervia(r_8,c(\miis,X,\midepressed),l). \\
  &\minumberofbodies(r_8,1).
\end{align*}
Axiom \eqref{eqEx4} \quo{$\miis(X,\midepressed) \lax \miis(Y,\midead) \lands \miimportantfor(Y,X)$}
is rewritten as follows.
\begin{align*}
  &\mimayinfervia(r_9,c(\miis,X,\midepressed),l(Y)) \lars
    \mipot(c(\miis,X,\midepressed)), Y \eqs s(r_9,``Y",X). \\
  &\miinferenceneeds(c(\miis,X,\midepressed),r_9,c(\miimportantfor,Y,X)) \lars \\
   &\eqhspacer \mimayinfervia(r_9,c(\miis,X,\midepressed),l(Y)). \\
  &\miinferenceneeds(c(\miis,X,\midepressed),r_9,c(\miis,Y,\midead)) \lars \\
   &\eqhspacer \mimayinfervia(r_9,c(\miis,X,\midepressed),l(Y)). \\
  &\minumberofbodies(r_9,2).
\end{align*}
Axiom \eqref{eqEx6} \quo{$\milost(X,Y) \lax \miis(Y,\midead) \lands \miimportantfor(Y,X) \lands \miinst(Y,\miperson)$}
is rewritten into the following ASP rules.
\begin{align*}
  &\mimayinfervia(r_{10},c(\milost,X,Y),l) \lars \mipot(c(\milost,X,Y)). \\
  &\miinferenceneeds(c(\milost,X,Y),r_{10},c(\miinst,Y,\miperson)) \lars 
   \mimayinfervia(r_{10},c(\milost,X,Y),l). \\
  &\miinferenceneeds(c(\milost,X,Y),r_{10},c(\miimportantfor,Y,X)) \lars 
   \mimayinfervia(r_{10},c(\milost,X,Y),l). \\ 
  &\miinferenceneeds(c(\milost,X,Y),r_{10},c(\miis,Y,\midead)) \lars
   \mimayinfervia(r_{10},c(\milost,X,Y),l). \\
  &\minumberofbodies(r_{10},3).
\end{align*}

%$comment("<- \miinst(X,\mimale), \miinst(X,\mifemale)").$ 
%$ \lars \mitrue(c(\miinst,MyX0,\mimale)), \mitrue(c(\miinst,MyX2,\mifemale)), eq(MyX0,MyX2).$
%$ comment("<- \mifatherof(X,Y), \miinst(X,\mifemale)").$ 
%$ \lars \mitrue(c(\mifatherof,MyX0,Y)), \mitrue(c(\miinst,MyX2,\mifemale)), eq(MyX0,MyX2).$

\subsection{Example Answer Set}

% terms: in real ASP, in Figure 1, in paper ASP
% s(f,1) = n_2 = s(r_4,``Y",f)
% s(s,2) = n_1 = s(r_9,``Y",s)

We next give parts of the answer set
representing the abductive explanation depicted in Figure~\ref{figExWA}.
We show an answer set of the encoding \bwdg\ which
does not perform any canonicalization and therefore
can produce the proof graph in Figure~\ref{figExWA}
(other encodings will produce larger proof graphs
with the same cost for this instance).

Note that the encoding produces Skolem terms
$s(r_9,``Y",s)$ and $s(r_4,``Y",f)$,
which, for practical reasons,
have been displayed in Figure~\ref{figExWA}
as $n_1$ and $n_2$, respectively.

\paragraph{Deterministically determined Atoms.}
Based on the rewriting of axioms (previous section)
and the goal atoms,
the truth of atoms of form $\mipot(\cdot)$,
$\mimayinfervia(\cdot,\cdot,\cdot)$, and
$\miinferenceneeds(\cdot,\cdot,\cdot)$,
is deterministically determined via rules
\eqref{eqPotential} and \eqref{eqNeedPotential}.

In our example we obtain the following true atoms in the answer set.
\begin{gather*}
\begin{align*}
  &\mipot(c(\mifatherof,f,s(r_4,``Y",f)))
    &&\mipot(c(\mifatherof,f,m)) \\
  &\mipot(c(\mifatherof,s(r_9,``Y",s),s))
   &&\mipot(c(\miimportantfor,f,m)) \\
  &\mipot(c(\miimportantfor,s(r_9,``Y",s),s))
   &&\mipot(c(\miinst,f,\mifemale)) \\
  &\mipot(c(\miinst,f,\mimale))
    &&\mipot(c(\miinst,f,\miperson))
    &&\mipot(c(\miinst,s,\mifemale)) \\
  &\mipot(c(\miis,s(r_9,``Y",s),\midead))
    &&\mipot(c(\miinst,s,\mipessimist))
    &&\mipot(c(\miis,f,\midead)) \\
  &\mipot(c(\miis,s,\midepressed))
    &&\mipot(c(\milost,m,f))
    &&\mipot(c(\miname,f,\mimary)) \\
  &\mipot(c(\miname,m,\mimary))
    &&\mipot(c(\miname,s,\mimary))
\end{align*} \\[1ex]
\begin{align*}
  &\mimayinfervia(r_4,c(\miinst,f,\mimale),l(s(r_4,``Y",f)))
    &&\mimayinfervia(r_5,c(\miinst,f,\mifemale),l) \\
  &\mimayinfervia(r_5,c(\miinst,s,\mifemale),l)
    &&\mimayinfervia(r_6,c(\miimportantfor,f,m),l) \\
  &\mimayinfervia(r_6,c(\miimportantfor,s(r_9,``Y",s),s),l)
    &&\mimayinfervia(r_7,c(\miinst,f,\miperson),l) \\
  &\mimayinfervia(r_8,c(\miis,s,\midepressed),l) \\
  &\mimayinfervia(r_9,c(\miis,s,\midepressed),l(s(r_9,``Y",s)))
    &&\mimayinfervia(r_{10},c(\milost,m,f),l)
\end{align*} \\[1ex]
\begin{align*}
  &\miinferenceneeds(c(\miinst,f,\mimale),r_4,c(\mifatherof,f,s(r_4,``Y",f))) \\
  &\miinferenceneeds(c(\miinst,f,\mifemale),r_5,c(\miname,f,\mimary)) \\
  &\miinferenceneeds(c(\miinst,s,\mifemale),r_5,c(\miname,s,\mimary)) \\
  &\miinferenceneeds(c(\miimportantfor,f,m),r_6,c(\mifatherof,f,m)) \\
  &\miinferenceneeds(c(\miimportantfor,s(r_9,``Y",s),s),r_6,c(\mifatherof,s(r_9,``Y",s),s)) \\
  &\miinferenceneeds(c(\miinst,f,\miperson),r_7,c(\miinst,f,\mimale)) \\
  &\miinferenceneeds(c(\miis,s,\midepressed),r_8,c(\miinst,s,\mipessimist)) \\
  &\miinferenceneeds(c(\miis,s,\midepressed),r_9,c(\miis,s(r_9,``Y",s),\midead)) \\
  &\miinferenceneeds(c(\miis,s,\midepressed),r_9,c(\miimportantfor,s(r_9,``Y",s),s)) \\
  &\miinferenceneeds(c(\milost,m,f),r_{10},c(\miinst,f,\miperson)) \\
  &\miinferenceneeds(c(\milost,m,f),r_{10},c(\miimportantfor,f,m)) \\
  &\miinferenceneeds(c(\milost,m,f),r_{10},c(\miis,f,\midead))
\end{align*}
\end{gather*}
From these atoms,
$\mihu(\cdot)$, and $\miuhu(\cdot)$
are deterministically determined using rules
\eqref{eqHu1}--\eqref{eqUhu}.
\begin{align*}
  &\mihu(m) 
    &&\mihu(\mifemale)
    &&\mihu(\miperson)
    &&\mihu(\midepressed)
    \\
  &\mihu(f) 
    &&\mihu(\mimale) 
    &&\mihu(\mipessimist)
    &&\mihu(s(r_4,``Y",f))
    \\
  &\mihu(s)
    &&\mihu(\mimary)
    &&\mihu(\midead)
    &&\mihu(s(r_9,``Y",s))
    \\
  &\miuhu(f)
  &&\miuhu(m)
  &&\miuhu(\mimary)
  &&\miuhu(\mipessimist) \\
  &\miuhu(s)
  &&\miuhu(s(r_4,``Y",f))
  &&\miuhu(s(r_9,``Y",s))
\end{align*}
Goal atoms are deterministically defined as true
by \eqref{eqGoalTrue}.
\begin{align*}
  &\mitrue(c(\miname,m,\mimary))
    && \mitrue(c(\miinst,s,\mifemale))
    && \mitrue(c(\milost,m,f)) \\
  &\mitrue(c(\mifatherof,f,m))
    &&\mitrue(c(\miis,s,\midepressed))
\end{align*}

\paragraph{Truth Justification.}
The \bwdg\ encodings requires a justification
for each true value,
this justification is nondeterministically guessed
to be either $\miinfer(\cdot)$, $\mifactor(\cdot)$,
or $\miabduce(\cdot)$ by rules
\eqref{eqTrueInferFai} and \eqref{eqFactorOrAbduce}.
Justifying an atom by inference
performs another nondeterministic guess
with \eqref{eqInferViaWhat}
about the concrete axiom used for that inference,
which is represented in $\miinfervia(\cdot,\cdot)$
and defines more atoms of form $\mitrue(\cdot)$ to be true
(and hence their need to be justified)
with \eqref{eqNeedInferTrue}.

In the answer set representing Figure~\ref{figExWA},
this guess contains the following true atoms
(we omit $\mifai(\cdot)$).
\begin{gather*}
\begin{align*}
  &\miinfer(c(\miinst,f,\mimale))
    &&\miinfervia(r_4,c(\miinst,f,\mimale)) \\
  &\miinfer(c(\miinst,s,\mifemale))
    &&\miinfervia(r_5,c(\miinst,s,\mifemale)) \\
  &\miinfer(c(\miimportantfor,f,m))
    &&\miinfervia(r_6,c(\miimportantfor,f,m)) \\
  &\miinfer(c(\miinst,f,\miperson))
    &&\miinfervia(r_7,c(\miinst,f,\miperson)) \\
  &\miinfer(c(\miis,s,\midepressed))
    &&\miinfervia(r_9,c(\miis,s,\midepressed)) \\
  &\miinfer(c(\milost,m,f))
    &&\miinfervia(r_{10},c(\milost,m,f)) \\
\end{align*} \\ % just for allowing page break
\begin{align*}
  &\mifactor(c(\miname,s,\mimary))
  &&\mifactor(c(\mifatherof,f,s(r_4,``Y",f))) \\
  &\mifactor(c(\miis,s(r_9,``Y",s),\midead))
  &&\mifactor(c(\miimportantfor,s(r_9,``Y",s),s)) \\
  &\miabduce(c(\miname,m,\mimary))
  &&\miabduce(c(\mifatherof,f,m)) \\
  &\miabduce(c(\miis,f,\midead)) \\
  &\mitrue(c(\miname,s,\mimary))
  &&\mitrue(c(\miinst,f,\miperson)) \\
  &\mitrue(c(\miinst,f,\mimale))
  &&\mitrue(c(\mifatherof,f,s(r_4,``Y",f))) \\
  &\mitrue(c(\miis,s(r_9,``Y",s),\midead))
  &&\mitrue(c(\miis,f,\midead)) \\
  &\mitrue(c(\miimportantfor,f,m))
  &&\mitrue(c(\miimportantfor,s(r_9,``Y",s),s))
\end{align*}
\end{gather*}

\paragraph{Term Equivalence.}
Independent from justification of true atoms,
an equivalence relation over all potential terms
is nondeterministically guessed by means of rules
\eqref{eqEqGuess}--\eqref{eqEqTransitive}.

The answer set representing Figure~\ref{figExWA}
contains the following true atoms for $\mieq(\cdot,\cdot)$
These atoms represent two equivalence classes
$\{m,s,s(r_4,``Y",f)\}$ and $\{f,s(r_9,``Y",s)\}$
that have more than one element,
and singleton equivalence classes
for sort names and other constants.
% TODO other constant singletons are not enforced!
\begin{align*}
  &\mieq(\mimale,\mimale)
    &&\mieq(\mifemale,\mifemale)
    &&\mieq(\midead,\midead) \\
  &\mieq(\mimary,\mimary)
    &&\mieq(\miperson,\miperson)
    &&\mieq(\mipessimist,\mipessimist) \\
  &\mieq(f,f)
    &&\mieq(f,s(r_9,``Y",s)
    &&\mieq(\midepressed,\midepressed) \\
  &\mieq(m,m)
    &&\mieq(m,s)
    &&\mieq(m,s(r_4,``Y",f)) \\
  &\mieq(s,m)
    &&\mieq(s,s)
    &&\mieq(s,s(r_4,``Y",f)) \\
  &\mieq(s(r_4,``Y",f),m)
    &&\mieq(s(r_4,``Y",f),s) 
    &&\mieq(s(r_4,``Y",f),s(r_4,``Y",f)) \\
  &\mieq(s(r_9,``Y",s),f)
    &&\mieq(s(r_9,``Y",s),s(r_9,``Y",s))
\end{align*}

\paragraph{Factoring.}
For each atom that was guessed as factored,
rules \eqref{eqGuessFactorInfer}--\eqref{eqFactorOk}
define $\mifactorvia(\cdot,\cdot)$
and $\mifactorOk(\cdot)$ for atoms that can be
factored with infered or abduced atoms
while respecting $\mieq(\cdot,\cdot)$ and acyclicity.
For acyclicity,
the partial order defined by the graph is represented in
$\mibelow(\cdot,\cdot)$, which is true
for every pair of atoms such that the first atom
is reachable via arcs from the second atom.
The representation of Figure~\ref{figExWA}
contains the following true atoms.
\begin{gather*}
\begin{align*}
  &\mifactorvia(c(\miname,s,\mimary),c(\miname,m,\mimary))   && \\
  &\mifactorvia(c(\miis,s(r_9,``Y",s),\midead),c(\miis,f,\midead)) \\
  &\mifactorvia(c(\miimportantfor,s(r_9,``Y",s),s),c(\miimportantfor,f,m)) \\
  &\mifactorvia(c(\mifatherof,f,s(r_4,``Y",f)),c(\mifatherof,f,m))
\end{align*} \\[1ex]
\begin{align*}
  &\mifactorOk(c(\miname,s,\mimary))
  &&\mifactorOk(c(\miis,s(r_9,``Y",s),\midead)) \\
  &\mifactorOk(c(\miimportantfor,s(r_9,``Y",s),s))
  &&\mifactorOk(c(\mifatherof,f,s(r_4,``Y",f))) \\
\end{align*} \\ % just for allowing page break
\begin{align*}
  &\mibelow(c(\miinst,s,\mifemale),c(\miname,m,\mimary))
  &&\mibelow(c(\miinst,s,\mifemale),c(\miname,s,\mimary)) \\
  &\mibelow(c(\miname,s,\mimary),c(\miname,m,\mimary))
  &&\mibelow(c(\milost,m,f),c(\miinst,f,\miperson)) \\
  &\mibelow(c(\milost,m,f),c(\miimportantfor,f,m))
  &&\mibelow(c(\milost,m,f),c(\miis,f,\midead)) \\
  &\mibelow(c(\milost,m,f),c(\miinst,f,\mimale))
  &&\mibelow(c(\milost,m,f),c(\mifatherof,f,m)) \\
  &\mibelow(c(\milost,m,f),c(\mifatherof,f,s(r_4,``Y",f)))
  &&\mibelow(c(\miinst,f,\miperson),c(\miinst,f,\mimale)) \\
  &\mibelow(c(\miinst,f,\miperson),c(\mifatherof,f,m))
  &&\mibelow(c(\miis,s,\midepressed),c(\miis,f,\midead)) \\
  &\mibelow(c(\miinst,f,\mimale),c(\mifatherof,f,s(r_4,``Y",f)))
  &&\mibelow(c(\miinst,f,\mimale),c(\mifatherof,f,m)) \\
  &\mibelow(c(\miis,s,\midepressed),c(\miimportantfor,f,m))
  &&\mibelow(c(\miis,s,\midepressed),c(\mifatherof,f,m)) \\
  &\mibelow(c(\miimportantfor,f,m),c(\mifatherof,f,m)) 
\end{align*} \\ %[1ex]
\begin{align*}
  &\mibelow(c(\mifatherof,f,s(r_4,``Y",f)),c(\mifatherof,f,m)) && \\
  &\mibelow(c(\miis,s(r_9,``Y",s),\midead),c(\miis,f,\midead)) \\
  &\mibelow(c(\miinst,f,\miperson),c(\mifatherof,f,s(r_4,``Y",f))) \\
  &\mibelow(c(\miis,s,\midepressed),c(\miis,s(r_9,``Y",s),\midead)) \\
  &\mibelow(c(\miis,s,\midepressed),c(\miimportantfor,s(r_9,``Y",s),s)) \\
  &\mibelow(c(\miimportantfor,s(r_9,``Y",s),s),c(\mifatherof,f,m)) \\
  &\mibelow(c(\miimportantfor,s(r_9,``Y",s),s),c(\miimportantfor,f,m))
\end{align*}
\end{gather*}
For representing the cost of the solution
wrt.\ objective \waobj,
\eqref{eqWAGoal} defines the following goal costs.
\begin{align*}
  &\mipcost(c(\miname,m,\mimary),100)
    &&\mipcost(c(\milost,m,f),100)
  &&\mipcost(c(\mifatherof,f,m),100) \\
  &\mipcost(c(\miinst,s,\mifemale),100)
    &&\mipcost(c(\miis,s,\midepressed),100)
\end{align*}
Cost propagation via inferences
is done by \eqref{eqWAInfer} which
makes the following atoms true.
\begin{align*}
  &\mipcost(c(\miname,s,\mimary),120)
    &&\mipcost(c(\miinst,f,\miperson),40) \\
  &\mipcost(c(\miimportantfor,f,m),40)
    &&\mipcost(c(\miis,f,\midead),40) \\
  &\mipcost(c(\miinst,f,\mimale),48)
    &&\mipcost(c(\mifatherof,f,s(r_4,``Y",f)),57) \\
  &\mipcost(c(\miis,s(r_9,``Y",s),\midead),60)
    &&\mipcost(c(\miimportantfor,s(r_9,``Y",s),s),60) \\
  &\mipcost(c(\mifatherof,f,m),48)
    &&\mipcost(c(\mifatherof,f,m),72)
\end{align*}
Note that the cost of $72\$$ is not shown in Figure~\ref{figExWA}
because it is not contained in the definition of
proof graph (Definition~\ref{defHypGraph}).
In the Proof of Proposition~\ref{thmObjectiveEncodings}
we argue that including these costs in ASP
only adds further (higher) costs to each atom,
hence the minimal costs
and therefore the optimal solution remain unchanged.
Note that we include these costs
(i.e., we omit one more minimization step)
for efficiency reasons.

Cost propagation via factoring is
realized in \eqref{eqWAFactor}
and makes the following atoms true.
\begin{align*}
  &\mipcost(c(\miname,m,\mimary),120)
    &&\mipcost(c(\mifatherof,f,m),57) \\
  &\mipcost(c(\miis,f,\midead),60)
    &&\mipcost(c(\miimportantfor,f,m),60)
\end{align*}
Actual cost is defined from potential cost
via \eqref{eqWACost}
which yields the following true atoms.
\begin{align*}
  &\micost(c(\miname,m,\mimary),100)
    &&\micost(c(\mifatherof,f,m),48)
    &&\micost(c(\miis,f,\midead),40)
\end{align*}
From these atoms, the constraint \eqref{eqWAObj}
obtains the overall cost of $188$ for this answer set.

}

\section{Appendix: Proofs for ASP Encoding Correctness}
\label{secProofs}
\begin{proofof}{Lemma~\ref{thmPbptProofGraph}}
  \myProofThmPbptProofGraph
\end{proofof}
\begin{proofof}{Proposition~\ref{thmBwdGuessProofGraph}}
  \myProofThmBwdGuessProofGraph
\end{proofof}
\begin{proofof}{Lemma~\ref{thmBwdGuessEq}}
  \myProofThmBwdGuessEq
\end{proofof}
\begin{proofof}{Proposition~\ref{thmBwda}}
  \myProofThmBwda
\end{proofof}
\begin{proofof}{Proposition~\ref{thmObjectiveEncodings}}
  \myProofThmObjectiveEncodings
\end{proofof}

\myfigureOnModelWithoutOptimization{tpb}

\section{Appendix: Realizing On-Demand Constraints with Optimization}
\label{secOnDemandAppendix}

If we solve a problem without weak constraints,
i.e., without optimization,
realizing on-demand constraints is simple:
we register the callback \textsc{OnModelWithoutOptimization}
(Algorithm~\ref{algoOnModelWithoutOptimization})
to \clasp.
This callback checks on-demand constraints in \textsc{FindOnDemandViolations}
(which is an application-specific algorithm),
adds nogoods for violated constraints,
and prints (or otherwise processes) the answer set
if no constraint was violated.
The first model we print this way is the first model
that does not violate on-demand constraints.

However, in the presence of optimization,
\clasp\ has two modes that are both unsuitable
with on-demand constraints:
in mode (opt) each answer set
updates the optimization bound
and subsequent answer sets must be better;
in mode (enum) we have to explicitly specify a bound
and answer set candidates of same or better quality
are found.
With (opt) it can happen that the first answer set
with optimal cost violates an on-demand constraint,
so we discard that model, but the solver will not find
further models with same cost,
but no models with better cost exist,
so we will not find any models (even if some exist).
With (enum) the search is blind as better models
are found only by chance.
A straightforward and more elegant solution would be,
to update the bound only for good answer sets in the on\_model callback,
but the API currently does not allow this.

To solve this problem we created algorithm
\textsc{FindOptimalModel} (Algorithm~\ref{algoFindOptimalModel})
for finding an optimal model with on-demand constraints.
This algorithm first grounds the program $P$ in line 1,
then uses (opt) mode
to find an optimistic bound ($\mi{optimisticCost}$)
for the cost of the optimal model in lines 2--4
using callback \textsc{OnModelFindBound}
(Algorithm~\ref{algoOnModelFindBound}).
This callback records the cost of the best encountered
model candidate and the best model that does not violate
on-demand constraints ($\mi{bestModel}$).
This search aggressively updates the bound
and also uses on-demand constraints.
If no optimistic cost is set,
the callback was never called and we return UNSAT (line 5).
If the cost of the best found feasible model
is the optimal cost, we directly return this model as optimal solution (line 6).
Otherwise, we enter a loop that enumerates models
using callback \textsc{OnModelFindSolution}
(Algorithm~\ref{algoOnModelFindSolution}).
The loop increases the optimization bound of the solver by one in each iteration,
until an answer set
that does not violate on-demand constraints can be found.
Our abduction instances always have some solution,
so we will find that solution and leave the endless loop
in line 13.
To make the algorithm terminate in the general case
where on-demand constraints might eliminate all solutions,
we can obtain the worst-case cost from the instantiation
of all weak constraints,
and abort the loop once we increment $\mi{tryingCost}$ to that cost.

\myfigureMainAlgorithm{bth}

\fi %

\end{document}

%% file: abstract.txt
We study abduction in First Order Horn logic theories
where all atoms can be abduced
and we are looking for preferred solutions
with respect to three objective functions:
cardinality minimality, coherence, and weighted abduction.
We represent this reasoning problem in Answer Set Programming (ASP),
in order to obtain a flexible framework for experimenting with
global constraints and objective functions,
and to test the boundaries of what is possible with ASP.
Realizing this problem in ASP is challenging 
as it requires value invention
and equivalence between certain constants, because
the Unique Names Assumption does not hold in general.
To permit reasoning in cyclic theories,
we formally describe fine-grained variations of limiting Skolemization.
We identify term equivalence as a main instantiation bottleneck,
and improve the efficiency of our approach
with on-demand constraints that were used to eliminate the same bottleneck
in state-of-the-art solvers.
We evaluate our approach experimentally on the ACCEL benchmark
for plan recognition in Natural Language Understanding.
Our encodings are publicly available, modular,
and our approach is more efficient than state-of-the-art solvers on the ACCEL benchmark.